\documentclass{article}
\usepackage{epsfig}
\usepackage{graphicx}
\usepackage{amsmath}
\usepackage[preprint]{spconfa4}
\usepackage{amssymb}
\usepackage{upgreek}
\usepackage{subfigure}
\usepackage{appendix}
\usepackage[ruled]{algorithm2e}
\usepackage{color}
\usepackage{frame}
\usepackage{multirow}

\graphicspath{{figures/}}
\newcommand{\tabincell}[2]{\begin{tabular}{@{}#1@{}}#2\end{tabular}}

\usepackage[pagebackref=true,breaklinks=true,colorlinks,bookmarks=false]{hyperref}

\begin{document}

\title{Fast and Interpretable 2D Homography Decomposition: Similarity-Kernel-Similarity and Affine-Core-Affine Transformations}

\name{Shen~Cai$^{\dagger,\ast}$,Zhanhao~Wu$^{\dagger}$,Lingxi~Guo$^{\dagger}$,Jiachun~Wang$^{\dagger}$,Siyu~Zhang$^{\dagger}$,Junchi~Yan$^{\ddagger}$,and~Shuhan~Shen$^{\S}$}

\address{$^{\dagger}$Visual and Geometric Perception Lab, Donghua University\\ 
$^{\ddagger}$Department of Computer Science and Engineering, Shanghai Jiao Tong University\\
$^{\S}$National Laboratory of Pattern Recognition, Institute of Automation, Chinese Academy of Sciences\\
$^{\ast}$Corresponding author: hammer\_cai@163.com}

\maketitle

\newcommand\blfootnote[1]{%
\begingroup 
\renewcommand\thefootnote{}\footnote{#1}%
\addtocounter{footnote}{-1}%
\endgroup 
}


\begin{abstract}

In this paper, we present two fast and interpretable decomposition methods for 2D homography, which are named Similarity-Kernel-Similarity (SKS) and Affine-Core-Affine (ACA) transformations respectively. Under the minimal $4$-point configuration, the first and the last similarity transformations in SKS are computed by two anchor points on target and source planes, respectively. Then, the other two point correspondences can be exploited to compute the middle kernel transformation with only four parameters. Furthermore, ACA uses three anchor points to compute the first and the last affine transformations, followed by computation of the middle core transformation utilizing the other one point correspondence. ACA can compute a homography up to a scale with only $85$ floating-point operations (FLOPs), without even any division operations. Therefore, as a plug-in module, ACA facilitates the traditional feature-based Random Sample Consensus (RANSAC) pipeline, as well as deep homography pipelines estimating $4$-point offsets. In addition to the advantages of geometric parameterization and computational efficiency, SKS and ACA can express each element of homography by a polynomial of input coordinates ($7$th degree to $9$th degree), extend the existing essential Similarity-Affine-Projective (SAP) decomposition and calculate 2D affine transformations in a unified way. Source codes are released in \url{https://github.com/cscvlab/SKS-Homography}.
\end{abstract}


\blfootnote{The work is partially supported by National Major State Research Development Program (2020AAA0107600), Shanghai Municipal Science and Technology Major Project (2021SHZDZX0102), and the Foundation of Key Laboratory of Artificial Intelligence, Ministry of Education, P.R. China (AI2020003).}

\section{Introduction}
\label{sec:intro}
\begin{figure}[t]
\begin{center}
\includegraphics[width=0.49\textwidth]{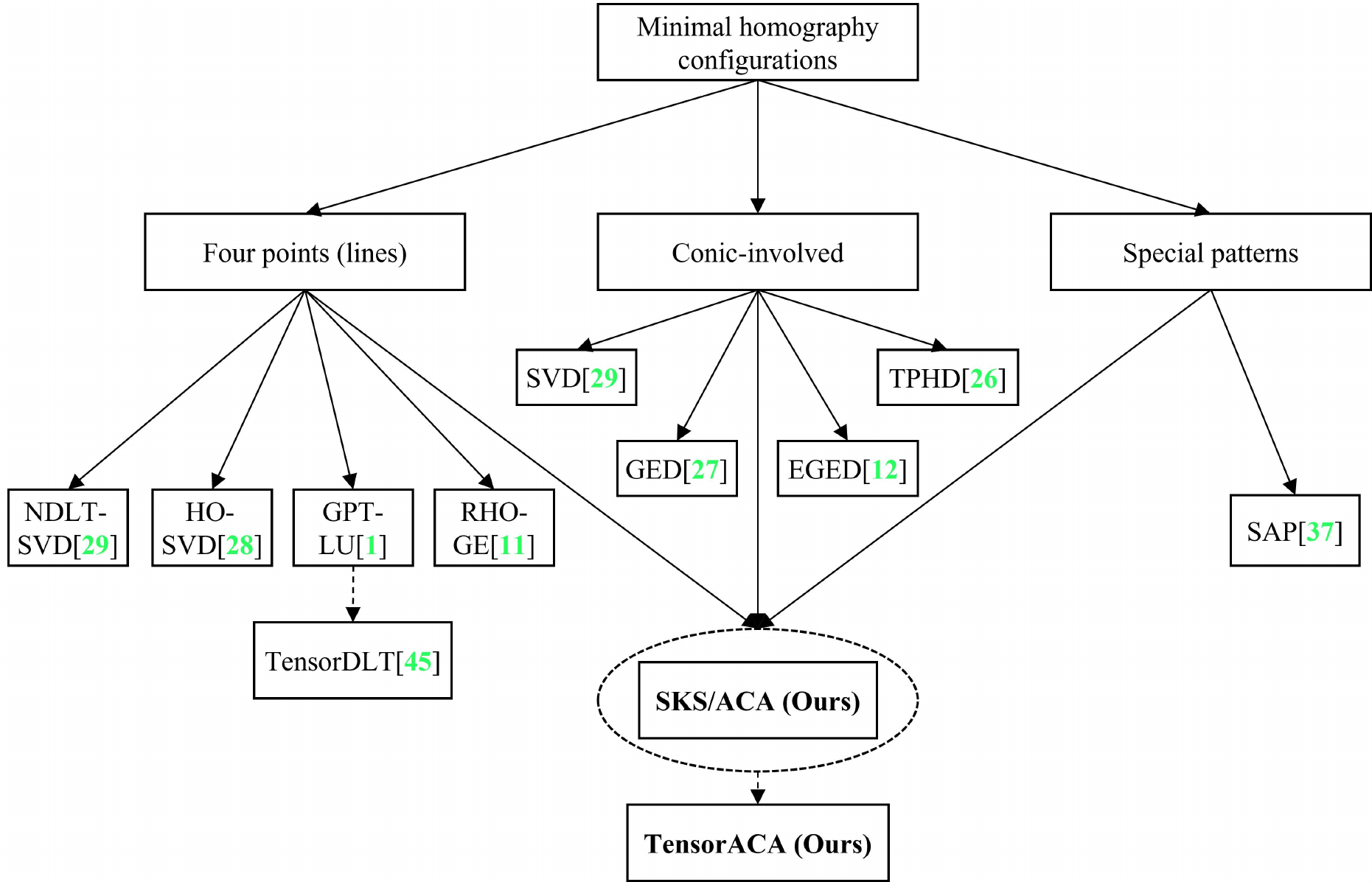}
\end{center}
\vspace{-5mm}
\caption{Sketch of homography computation methods under the minimal condition. According to various primitive configurations and underlying mathematical principles, solving methods may differ considerably. The proposed SKS and ACA can deal with three categories of primitive configurations in a unified way, while is suitable to be tensorized in deep homography pipeline.} 
\vspace{-5mm}
\label{fig:homography_overview}
\end{figure}

Planar homography, also referred as two-dimensional (2D) projective transformation, plays an important role in many tasks of geometric vision, such as camera calibration~\cite{Zhang_PAMI00}, plane based pose estimation~\cite{IPPE2014}, image stitching~\cite{ImageStitch_CVPR13} and monocular motion estimation~\cite{ORB-SLAM3}. A 2D homography denoted by a $3\!*\!3$ homogeneous matrix has $8$ degrees of freedom (DOF). 
Assuming coplanar geometric primitives (such as points, lines and conics) are given, a homography may be computed under the minimal condition or under the over-determined condition, depending on DOF of primitives. In this paper, we focus on the homography computation problem under the minimal condition. Previous approaches explore various algebraic or geometric ways to study the minimal planar configurations. According to the categories of geometric primitives, they could be roughly divided into three categories: i) four points or lines; ii) conic-involved patterns; iii) special patterns used in the hierarchical homography decomposition. Please see Fig.~\ref{fig:homography_overview} for a brief overview.

\textbf{i) Four points or lines}. Among all planar minimal configurations, $4$-point pattern is the most common partly due to the mature interest points extraction and matching algorithms, such as traditional SIFT~\cite{SIFT2004}, SURF~\cite{SURF2006}, ORB~\cite{ORB} and deep learning based LIFT~\cite{LIFT_ECCV16}, SOSNet~\cite{SOSNet_CVPR19}, SuperPoint~\cite{SuperPoint18}.
Moreover, predicting $4$-point offsets in target image (while fixing $4$-point in source image) is also popular in deep homography estimation algorithms~\cite{UDHN_RAL18, CAUDHN_ECCV20, IDHN_CVPR22, UDIS_TIP21, DHDS_CVPR20, LocalTrans_ICCV21, DAMG_TCSVT22} as this parameterization of homography~\cite{Kanade_TechnicalReport06} is meaningful and beneficial for training neural networks~\cite{arXiv16}.

\begin{figure*}[th]
\begin{center}
\includegraphics[width=0.8\textwidth]{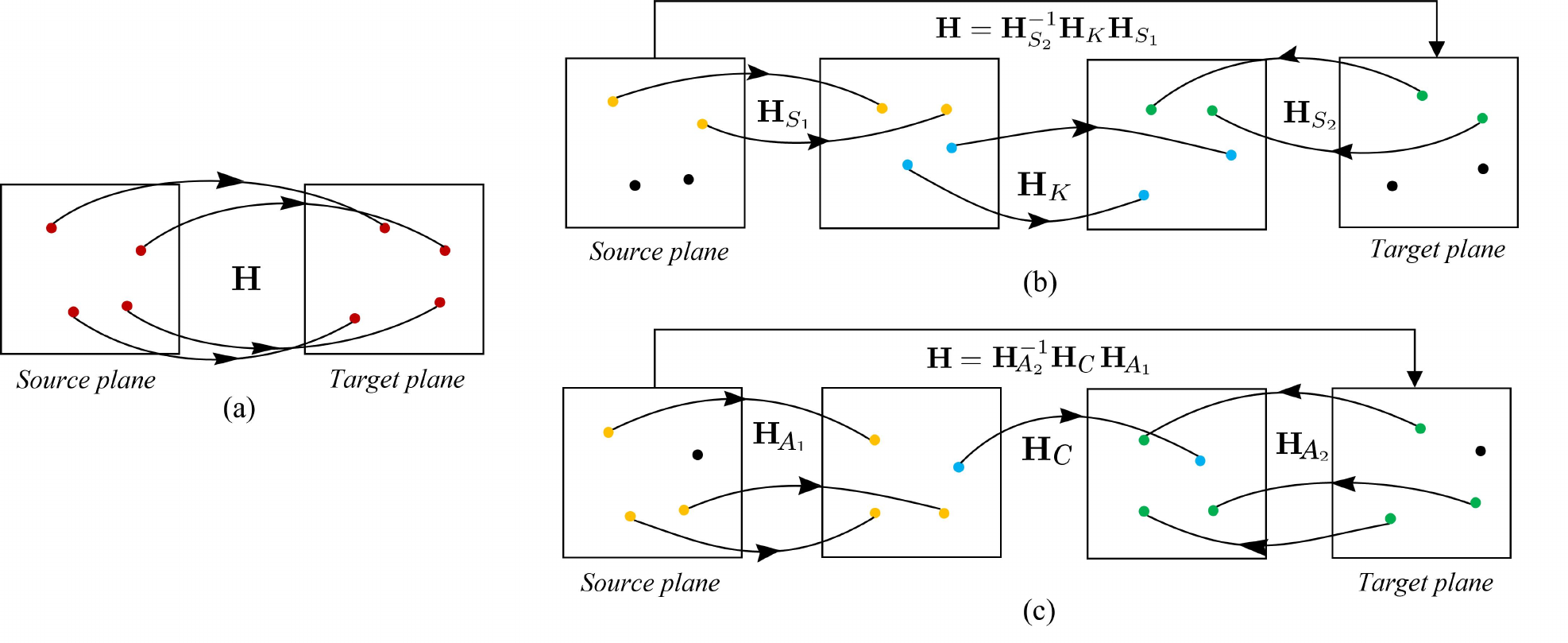}
\end{center}
\vspace{-7mm}
\caption{Comparison of solving process between previous $4$-point homography methods and ours. (a) The previous $4$-point homography methods (NDLT-SVD~\cite{Hartley2003Multiple}, HO-SVD~\cite{HO_BMVC05}, GPU-LU~\cite{OpenCV_GPT}, RHO-GE~\cite{Bazargani2015Fast}) indiscriminately utilize all four point correspondences (drawn with red dots) to construct a linear system (Eq.~\ref{equ:intro_3}) and solves it using well-established matrix factorization methods. 
(b) The proposed SKS decomposition includes three stratified sub-transformations, which are the first similarity transformation $\mathbf{H}_{S_1}$, the middle kernel transformation $\mathbf{H}_{K}$ and the last similarity transformation $\mathbf{H}_{S_2}$. Two corresponding points on source plane (drawn with yellow) and target plane (drawn with green) are used to compute $\mathbf{H}_{S_1}$ and $\mathbf{H}_{S_2}$ respectively. The residual two point correspondences (drawn with blue) on the two intermediate planes are used to compute $\mathbf{H}_{K}$ with $4$ DOF. (c) The proposed ACA decomposition is similar to SKS, except three point correspondence is used to compute the affine transformation $\mathbf{H}_{A_1}$ and $\mathbf{H}_{A_2}$, while one point correspondence is used to compute the core transformation $\mathbf{H}_{C}$ with $2$ DOF.}
\vspace{-2mm}
\label{fig:intro_realtrans}
\end{figure*}

Previous $4$-point homography computation methods firstly construct a square system of linear equations, followed by adopting different mathematical solvers to improve computational efficiency for the constructed linear system. 
The classical homography computation method is the normalized direct linear transform (NDLT)~\cite{Hartley2003Multiple}~[p.~109], which normalizes data first and then adopts singular value decomposition (SVD)~\cite{Golub2013}~[p.~111] to solve the formed $9$-variable linear system. This method is called NDLT-SVD in this paper to highlight its mathematical solver. The HO-SVD method~\cite{HO_BMVC05} utilizes the perpendicular properties of columns in coefficient matrix to simplify the $9$-variable linear system to a $3$-variable style, which significantly reduces the computation amount of SVD without loss of robustness. However, the above two methods simultaneously dealing with $4$-point and $n$-point ($n$$>$$4$) patterns contain many redundant operations when solving square linear systems. Specially designed for the $4$-point pattern, GPT-LU~\cite{OpenCV_GPT} forms an square linear system of equations and adopts LU factorization~\cite{Golub2013}~[p.~114] to solve it. RHO-GE~\cite{Bazargani2015Fast} designs a customized Gaussian elimination (GE)~\cite{Golub2013}~[p.~111] algorithm for the coefficient matrix $\mathbf{A}$ containing a large number of $0$ elements, which further accelerates the $4$-point homography computation. However, all the above $4$-point homography computation methods are limited by the strategy of constructing the square linear system first, which exists a lot of redundant operations. As shown in Fig.~\ref{fig:intro_realtrans} (a), for these four
methods, four point correspondences (drawn with red dots) are indiscriminately utilized to construct a linear system where the entire homography is unknown.

The above mentioned $4$-point homography methods are widely applied in traditional random sample consensus (RANSAC)~\cite{fischler_ACMC81} framework and its variants (e.g., USAC~\cite{Raguram2013USAC} and MAGSAC++~\cite{MAGSACplusplus}) to estimate homography between two images with outliers. In end-to-end deep homography methods predicting the offsets of four corner points, the adopted solver is the tensorized GPT-LU (firstly named TensorDLT in~\cite{UDHN_RAL18}). 

For line primitives which may be extracted by traditional algorithms~\cite{HoughTransform, LSD} or deep learning algorithms~\cite{DeepHT, SOLD2}, it is proved that $4$-line pattern is geometrically dual to $4$-point pattern. Therefore, all the above $4$-point homography methods can be directly adopted for the $4$-line pattern. 

\textbf{ii) Conic-involved patterns}. Conic-involved patterns mainly include two conics or one conic with coplanar points (lines). There already exist many works to study patterns with a pair of coplanar conics, such as concentric circles~\cite{ICCV05}, coplanar circles~\cite{Wu_IVC06, Chen_ECCV04}, confocal conics~\cite{Gurdjos_CVPR06} and coplanar conics~\cite{Mine_MVA12}.
Moreover, the conic-line(point) hybrid configuration is also studied for homography computation and camera calibration~\cite{Guo2017A}.
These methods generally build the link between the general intersection points of conics and one special kind of imaginary points, such as the circular points~\cite{Projective1952}~[p.~33],~\cite{Hartley2003Multiple}~[p.~53], to avoid solving the nonlinear constraints formed by projection equality of one conic correspondence.
However, the computational efficiency of these methods is not high as the time-consuming operations, such as SVD, are needed.
Furthermore, the geometric meaning of the matrices in these methods is limited without explanation of each parameter.

\textbf{iii) Special patterns used in the hierarchical homography decomposition}.
There exists another well-known homography decomposition method~\cite{liebowitz_CVPR98},~\cite{Hartley2003Multiple}~[p.~42-43], which express a projective transformation into a geometrical hierarchy chain of essential similarity, affine, and projective (SAP) transformations. 
The pattern used to hierarchically recover affine and metric properties should include two parts consisting of specific primitives. 
The first part~\cite{Hartley2003Multiple}~[p.~49-51] consists of two sets of three collinear points with known length ratios or two two sets of parallel lines, which are used to obtain the vanishing line and compute the essential projective transformation.
The second part~\cite{Hartley2003Multiple}~[p.~56-57] consists of two orthogonal line pairs, which are used to constrain the conic dual to the circular points to compute the affine transformation.
However, this decomposition method is almost never used in practice, mainly because its decomposition chain cannot be computed from general primitives, such as unconstrained points or lines.

The above three categories of homography computation methods under the minimal condition have their own drawbacks. The geometric homography decomposition methods require special primitives configurations or have high computational complexity. The algebraic $4$-point homography methods lack geometric meaning and are not fully explored in computation efficiency.
A desirable homography computation method preferably has the following properties:

 \vspace{1mm}
    i) Different planar primitives can be handled in a unified way.
    
    ii) Parameters involved have clear geometric meanings.
    
    iii) Homography is computed at a superior efficiency.
    
    iv) Solving process is suitable for being tensorized. 
    \vspace{1mm}

In this paper, a new homography decomposition formula with stratified geometric meaning and high computation efficiency is proposed to deal with the common four point correspondences.
This $4$-point homography computation method enjoys all aforementioned desirable properties.
Specifically, a homography is decomposed into three sub-transformations as shown in Fig.~\ref{fig:intro_realtrans} (b), which are \emph{similarity-kernel-similarity} (SKS) transformations respectively.
Two point correspondences as two anchor points (TAP) drawn with yellow dots on source plane and drawn with green dots on target plane are utilized to compute the last and first similarity transformations $\mathbf{H}_{S_1}$ and $\mathbf{H}_{S_2}$, respectively.
The similarity transformation plays the role of normalizing TAP to a canonical form.
After transforming the two planes under $\mathbf{H}_{S_1}$ and $\mathbf{H}_{S_2}$ respectively, the other two point correspondences drawn with blue can be used to linearly solve the kernel transformation $\mathbf{H}_{K}$ with only $4$ parameters.
Since the complicated and time-consuming operations, such as singular value decomposition (SVD), are avoided, the calculation amount for each step of homography computation is only a few dozens of floating-point operations (FLOPs).
Moreover, an improved \emph{affine-core-affine} (ACA) decomposition shown in Fig.~\ref{fig:intro_realtrans} (c) is also derived to further accelerate homography computation. Three anchor points (HAP) in source and target planes are utilized to compute the the last and first affine transformations $\mathbf{H}_{A_1}$ and $\mathbf{H}_{A_2}$, respectively. The transformed fourth point correspondence is used to solve the middle core transformation $\mathbf{H}_{C}$ with only $2$ parameters. 

The main contributions of this paper are concluded as follows:

    i) We propose the similarity-kernel-similarity (SKS) and affine-core-affine (ACA) transformations for 2D homography decomposition, with a variety of matrix factorization forms. We also prove that the existing SAP decomposition is just one special case of our formulas and 2D affine transformations can be computed in a unified way. 

    ii) In geometry, each sub-transformation of SKS or ACA, including each parameter in them, has a clear meaning. The first and last transformations in SKS or ACA symmetrically normalize target and source planes respectively, while the middle transformation determine the projective distortion between two normalized planes.
    
    iii) In algebra, SKS is superior to previous $4$-point homography methods in computational efficiency. Furthermore, ACA requires only $97$ FLOPs ($85$ FLOPs and no division operations for homographies up to a scale), which is about $5\%$ and $0.4\%$ of the state-of-the-art (SOTA) methods GPT-LU and NDLT-SVD respectively.
   In addition, we directly derive the polynomial expression of $16$ coordinates of four point correspondences for each element of homography.
    
    iv) 
    SKS or ACA can be integrated into traditional feature-based RANSAC pipelines to replace their default $4$-point homography solvers when estimating homography with outliers. At the same time, the tensorized ACA (TensorACA) is also suitable for integration into deep homography pipelines due to its concise homography expression and a small number of vector operations.

The rest of this paper is organized as follows. Section~\ref{sec:relatedWorks} reviews the related works of homography computation.
Section~\ref{sec:4Points} presents the derivation of SKS for $4$-point pattern.
Section~\ref{sec:ACA} illustrates the improved ACA homography decomposition. Section~\ref{sec:application} summaries the applications of our method in extension of SAP decomposition, decomposition for 2D affine transformation and tensorization for deep homography pipeline.
Section~\ref{sec:experiments} shows the experimental results. Section~\ref{sec:conclusion} gives the conclusion and future work.

\vspace{-2mm}
\section{Problem Formulation and Related Works}
\label{sec:relatedWorks}
\vspace{-1mm}
The proposed method mainly involves two sub-directions related to homography, which are points based homography computation and homography decomposition, assuming that corresponding coplanar primitives are given. These two sub-directions are formulated and reviewed in Sec.~\ref{sec:relatedWorks_LinearSystem} and Sec.~\ref{sec:relatedWorks_Decom}, respectively.
Meanwhile, $4$-point homography computation algorithms are generally called as a standalone module in the methods of estimating homography between two images. Thus, the traditional feature-based RANSAC pipelines and deep learning based homography estimation are reviewed in Sec.~\ref{sec:relatedWorks_RANSAC} and Sec.~\ref{sec:relatedWorks_DeepLearning}, respectively. 
Sec.~\ref{sec:relatedWorks_FLOPs} reviews methods evaluating computational amount of algorithms.
Table~\ref{tab:relatedWork_pointsHomography} and Table~\ref{tab:relatedWork_decomMethods} summarize the main differences between the proposed methods and the existing homography computation methods to provide an overall understanding and context.

\vspace{-2mm}
\subsection{Points based Homography Computation}
\label{sec:relatedWorks_LinearSystem}
Points are the most common geometric primitives.
Previous $4$-point homography methods~\cite{Hartley2003Multiple,HO_BMVC05,OpenCV_GPT,Bazargani2015Fast}  follow the same way to construct a square system of linear equations, whose first step is to remove the homogeneous equality of one point correspondence by utilizing the cross-product, i.e., 
\begin{equation}
    {\mathbf{x}}_2\!\equiv\! \mathbf{H}\mathbf{x}_1  \, \Rightarrow \,  \mathbf{x}_2\!\times\!(\mathbf{H}\mathbf{x}_1)\!=\!0,
    \label{equ:intro_1}
\end{equation}
where $\mathbf{H}$ denotes a $3$$*$$3$ homography matrix; $\mathbf{x}_1$ and $\mathbf{x}_2$ are $3$$*$$1$ homogeneous vectors of one point correspondence $\{\!{X_1}\!\!\stackrel{\mathbf{H}}\longrightarrow\!\!{X_2}\!\}$; `$\equiv$' denotes the equality up to a scale.

After some simplifications, two linear equations of homography are obtained,
\begin{equation}
    \begin{bmatrix}
    \mathbf{x}_1^{\mathrm{T}} & \mathbf{0}^{\mathrm{T}} & -X_2.x*\mathbf{x}_1^{\mathrm{T}} \\
     \mathbf{0}^{\mathrm{T}} & -\mathbf{x}_1^{\mathrm{T}} & X_2.y*\mathbf{x}_1^{\mathrm{T}}
    \end{bmatrix} \mathbf{h}=0,
    \label{equ:intro_2}
\end{equation}
where the suffixes $.x$ and $.y$ denote the coordinates of one 2D point; $\mathbf{h}$ is the vector form of homography. 

Stacking the eight linear equations provided by four point correspondences, a linear system of equations is constructed, 
\begin{equation}
\mathbf{A}_{8*9}*\mathbf{h}_{9*1}=0,
 \label{equ:intro_3}
\end{equation}
where $\mathbf{A}$ denotes the coefficient matrix.

\begin{table*}[t!]
\begin{center}
\centering
\caption{Comparison of $4$-point homography computation methods. Different from the previous four methods (shown in the second to fifth row), the proposed SKS and ACA do not need to form a linear system $\mathbf{A}\!*\!\mathbf{h}\!=\!0/\mathbf{b}$ and thus are denoted by $\bigotimes$. The notation $\S$ means that FLOPs is exactly counted from each arithmetical operation in source codes written in \textsc{C++} (where no external functions is called), while $\ge$ and $\sim$ denote an estimated lower bound and approximate value of FLOPs respectively. Among all methods, RHO-GE is not robust as pivoting is ignored for rapid calculation. In traditional feature-based pipelines, the previous four methods are integrated as $4$-point homography solver. In end-to-end deep homography pipelines, only the tensorized GPT-LU (called TensorDLT by~\cite{UDHN_RAL18}) is adopted to compute homography from $4$-point offsets.}
\label{tab:relatedWork_pointsHomography}
\vspace{3mm}
\resizebox{0.99\textwidth}{!}{
\begin{tabular}{|l|c|c|c|c|c|c|c|}
\hline
 1)\; \textbf{Method} & {\textbf{Formed Linear System}} &  {\textbf{Solving Way}} & {\textbf{FLOPs}} & {\textbf{Robustness}} &  {\textbf{Traditional Pipeline}} & {\textbf{End-to-End Deep Pipeline}}  \\
\hline
2)\; NDLT-SVD~\cite{Hartley2003Multiple} &  $\mathbf{A}_{8*9}*\mathbf{h}_{9*1}=0$  & algebraic & $\geq\!27400$ & \checkmark & \cite{Hartley2003Multiple,Raguram2013USAC,ORB-SLAM3} & / \\
 \hline
3)\; HO-SVD~\cite{HO_BMVC05} &  $\mathbf{A}_{8*3}*\mathbf{h}_{3*1}=0$  & algebraic & $\geq\!1800$ & \checkmark & \cite{IPPE2014} & / \\ 
\hline
\tabincell{l}{4)\; GPT-LU\cite{OpenCV_GPT} or \\ \;\; \; TensorDLT\cite{UDHN_RAL18}} &  $\mathbf{A}_{8*8}*\mathbf{h}_{8*1}=\mathbf{b}_{8*1}$  & algebraic &  $\sim\!1950$ & \checkmark & \cite{Raguram2013USAC,MAGSACplusplus} & \tabincell{c}{\cite{UDIS_TIP21, DHDS_CVPR20, CAUDHN_ECCV20}, \\  \cite{UDHN_RAL18, LocalTrans_ICCV21, DAMG_TCSVT22, IDHN_CVPR22} } \\
  \hline  
  5)\; RHO-GE~\cite{Bazargani2015Fast} &  $\mathbf{A}_{8*9}*\mathbf{h}_{9*1}=0$  & algebraic &  $223^\S$ & $\times$ & \cite{Bazargani2015Fast} & / \\
 \hline
 6)\; SKS (\textbf{Ours}) &  $\bigotimes$  & \tabincell{c}{ algebraic \& \\ geometric} & $169^\S$ & \checkmark & / & / \\
 \hline
 7)\; ACA (\textbf{Ours}) &  $\bigotimes$  & \tabincell{c}{ algebraic \& \\ geometric} & $97^\S$ & \checkmark & / & / \\
 \hline 
\end{tabular}
}
\vspace{-5mm}
\end{center}
\end{table*}

For the constructed linear system in Eq.~\ref{equ:intro_3}, previous methods may change its form and apply different matrix factorization methods to solve it. For example, NDLT-SVD~\cite{Hartley2003Multiple}~[p.~109] first normalizes all points on source and target planes and then applies SVD to solve the formed linear system under the constraint $||\mathbf{h}||$$=$$1$.
HO-SVD~\cite{HO_BMVC05} takes advantage of the sparse nature of the coefficient matrix to simplify the linear system representation to three unknowns. Thus the linear system for $n$ points is simplified to $\mathbf{A}_{2n*3}\!*\!\mathbf{h}_{3*1}\!=\!\!0$, which is adopted by the SOTA plane-based pose estimation method~\cite{IPPE2014} to accelerate homography computation. 
The above two methods can simultaneously deal with $4$-point pattern and $n$-point ($n$$>$$4$) pattern. As a result, the computational amount (FLOPs) of the above two methods for $4$-point pattern is quite large, as shown in the second and third rows in Table~\ref{tab:relatedWork_pointsHomography}. Since SVD is inherently iterative, we here give an estimated lower bound of FLOPs based on the OpenCV's implementation.

Specially designed for $4$-point pattern, the OpenCV's function \textit{getPerspectiveTransform}~\cite{OpenCV_GPT} (called GPT-LU here) converts the initial linear system to an inhomogeneous form: $\mathbf{A}_{8*8}$$*$$\mathbf{h}_{8*1}\!\!=\!\!\mathbf{b}_{8*1}$. Then this inhomogeneous linear system is solved by LU factorization~\cite{Golub2013}~[p.~114] which internally contains Gaussian elimination (GE) with partial pivoting\footnote{There are also other ways to solve this inhomogeneous linear system in various implementations, such as directly inverting $\mathbf{A}$ and utilizing QR decomposition. Here we choose the most common LU factorization with partial pivoting as a representative for this category of method.}.
Another fast algorithm specifically for $4$-point pattern is
RHO-GE~\cite{Bazargani2015Fast}, which utilizes the characteristics of the initial coefficient matrix $\mathbf{A}_{8*9}$ (containing a number of $0$ and $1$ elements) to simplify the process of GE. However, since this customized algorithm gets rid of choosing pivot element, this strategy may suffer from numerical robustness issues and is not recommended for use in linear algebra libraries, such as LAPACK and Eigen.
Although these two specialized $4$-point homography methods (shown in the fourth and fifth rows in Table~\ref{tab:relatedWork_pointsHomography}) improve the computational efficiency, both of them are still limited by the redundancy of the constructed linear system.

All the previous $4$-point homography methods shown in Table~\ref{tab:relatedWork_pointsHomography} follow Eq.~\ref{equ:intro_3} to construct a linear system. Consequently, the linear system can be solved by using a well-established matrix factorization method.
Totally different from the previous $4$-point homography methods, both the proposed SKS and ACA directly compute the sub-transformations in homography. Therefore, they are not related to the constructed linear system of equations and involve little computation. Moreover, SKS and ACA enjoy geometric meaning in each decomposition matrix, even in each parameter, as well as other algebraic advantages revealed later. 

\vspace{-2mm}
\subsection{Traditional Feature-based Pipelines}
\label{sec:relatedWorks_RANSAC}
Traditional pipelines of estimating homography between two images often include two stages.
In first stage, feature points are extracted by traditional methods~\cite{SIFT2004,SURF2006,ORB} or deep learning based methods~\cite{LIFT_ECCV16,SuperPoint18,SOSNet_CVPR19}, followed by coarse descriptor matching through brute force search. In second stage, outliers are removed and final homography is obtained under the RANSAC scheme~\cite{fischler_ACMC81} or its variants, such as PROSAC~\cite{PROSAC}, USAC~\cite{Raguram2013USAC}, MAGSAC~\cite{MAGSAC} and MAGSAC++~\cite{MAGSACplusplus}.
These traditional homography estimation pipelines differ from several aspects, of which this paper focuses on $4$-point homography solver.

The second column from right to left in Table~\ref{tab:relatedWork_pointsHomography} summaries the $4$-point homography algorithms in existing traditional pipelines. Specifically, NDLT-SVD is used in the standard RANSAC~\cite{Hartley2003Multiple}~[p.~123] implemented in OpenCV library, official implementation of USAC and the optimization-free RANSAC in official implementation of ORB-SLAM3~\cite{ORB-SLAM3}.
HO-SVD is used in the SOTA plane-based pose estimation method~\cite{IPPE2014}.
GPT-LU is used in USAC and MAGSAC++ implemented in OpenCV library.
RHO-GE is proposed in the work~\cite{Bazargani2015Fast} mainly based on PROSAC.
Due to the circular hypothesis and verifying scheme of RANSAC, the speed of $4$-point homography computation becomes non-negligible for the whole process, especially facing high outlier ratios.
The proposed SKS or ACA algorithm as a standalone module can be plugged into any of the above traditional pipelines to greatly reduce the runtime of $4$-point homography computation step, as well as the running time of whole process.
\begin{table*}[htb]
\begin{center}
\centering
\caption{Overview of geometry-based homography decomposition. $\bigotimes$ denotes the SAP method is not based on anchor points (AP) and does not have AP related FLOPs. $\dag$ denotes that all previous AP based methods use SVD to obtain decomposed matrices (refer to~\cite{Golub2013}~[p.~298] for a theoretical FLOPs of SVD). 
Our SKS and ACA avoid SVD and their FLOPs drop dramatically (the details are illustrated in Sec.~\ref{sec:4pts_FLOPs} and Sec.~\ref{sec:ACA_FLOPs} respectively.).}
\label{tab:relatedWork_decomMethods}
\vspace{3mm}
\resizebox{0.99\textwidth}{!}{
\begin{tabular}{|l|c|c|c|c|}
\hline  
 \tabincell{c}{ 1)\; \textbf{Method} } & \tabincell{c}{ \textbf{Pattern Configuration}} & \tabincell{c}{\textbf{Anchor Points (AP)}} & \tabincell{c}{\textbf{Decomposition Formula}} & \textbf{AP related FLOPs} \\
\hline 
\tabincell{c}{ 2)\; SVD~\cite{Hartley2003Multiple} } &
 \tabincell{c}{ 1. two circles \\
2. one circle and the infinity line } &
 \tabincell{c}{two circular points:  \\
 $\{I,J\}$} &
 \tabincell{c}{[$\tilde{I}\tilde{J}$]=$\mathbf{U}[{I}{J}]\mathbf{U}^\top$ \\ $\mathbf{H}=\mathbf{U}\mathbf{H}_{S}$} & $12\!*\!3^3\!=\!324^\dag$ \\
 \hline  
 \tabincell{c}{ 3)\; GED~\cite{Gurdjos_CVPR06} } &
 \tabincell{c}{ two confocal conics } 
& \tabincell{c}{ two circular points:  \\
 $\{I,J\}$ } &

\tabincell{c}{$\tilde{\mathbf{C}}_{1} - \lambda_{min}\tilde{\mathbf{C}}_{2} =[\tilde{I}\tilde{J}]=\mathbf{U}[{I}{J}]\mathbf{U}^\top$ \\ $\mathbf{H}=\mathbf{U}\mathbf{H}_{S}$} & $12\!*\!3^3\!=\!324 ^\dag$ \\
  \hline  
 \tabincell{c}{ 4)\; EGED~\cite{Mine_MVA12} } &
 \tabincell{c}{ two general conics } 
 & \tabincell{c}{ two complex conjugate points:  \\
$\{M,N\}$ } &
\tabincell{c}{$[{M}{N}]=\mathbf{H}^{-1}_{P}[{I}{J}]\mathbf{H}_{P}^{-\top}$ \\
$[\tilde{M}\tilde{N}]=\mathbf{H}_{Q}^{-1}[{I}{J}]\mathbf{H}_{Q}^{-\top}$ \\
$\mathbf{H} = \mathbf{H}_{Q}^{-1}\mathbf{H}_{S}\mathbf{H}_{P}$} & $2\!*\!12\!*\!3^3\!=\!648^\dag$ \\
\hline
 \tabincell{c}{ 5)\; TPHD~\cite{Guo2017A} } &
 \tabincell{c}{ 1. one conic and one line
\\ 2. four 2D points } &
 \tabincell{c}{ two real points: \\
 $\{M,N\}$ } &
 \tabincell{c}{$[{M}{N}]=\mathbf{H}_{P}^{-1}[{I}^{'}{J}^{'}]\mathbf{H}_{P}^{-\top}$  \\
 $[\tilde{M}\tilde{N}]=\mathbf{H}_{Q}^{-1}[{I}^{'}{J}^{'}]\mathbf{H}_{Q}^{-\top}$ \\
 $\mathbf{H} = \mathbf{H}_{Q}^{-1}\mathbf{H}^{'}_{S}\mathbf{H}_{P}$}  & $2\!*\!12\!*\!3^3\!=\!648^\dag$
 \\   \hline
 \tabincell{c}{ 6)\; SAP~\cite{liebowitz_CVPR98} } &
 \tabincell{c}{ the infinity line and  \\ two sets of orthogonal lines } &
 $\bigotimes$ &
 \tabincell{c}{ $\mathbf{H} = \mathbf{E}_{S}\mathbf{E}_A\mathbf{E}_{P}$} & $\bigotimes$ 
  \\    \hline
 \tabincell{c}{ 7)\; SKS (\textbf{Ours}) } &
 \tabincell{c}{ four 2D points } &
 \tabincell{c}{ two real points:  \\
$\{M,N\}$ } &
 \tabincell{c}{ $\{M,N\}{\stackrel{\mathbf{H}_{S_1}}{\longrightarrow}}[\mp1,0,1]^\top$ \\ 
 $\{\tilde{M},\tilde{N}\}{\stackrel{\mathbf{H}_{S_2}}{\longrightarrow}}[\mp1,0,1]^\top$ \\ 
 $\mathbf{H} = \mathbf{H}_{S_2}^{-1}\mathbf{H}_K\mathbf{H}_{S_1}$} & $2\!*\!9\!=\!18$  \\
 \hline
 \tabincell{c}{ 8)\; ACA (\textbf{Ours}) } &
 \tabincell{c}{ four 2D points } &
 \tabincell{c}{ three real points:  \\
$\{M,N,P\}$ } &
 \tabincell{c}{ $\{M,N,P\}{\stackrel{\mathbf{H}_{A_1}}{\longrightarrow}}[0,0,1]^\top, [0,1,1]^\top,[1,0,1]^\top$ 
 \\ $\{\tilde{M},\tilde{N},\tilde{P}\}{\stackrel{\mathbf{H}_{A_2}}{\longrightarrow}}[0,0,1]^\top, [0,1,1]^\top,[1,0,1]^\top$ 
 \\ $\mathbf{H} = \mathbf{H}_{A_2}^{-1}\mathbf{H}_C\mathbf{H}_{A_1}$} & $2\!*\!7\!=\!14$  \\
 \hline 
\end{tabular}
}
\vspace{-4mm}
\end{center}
\end{table*}

\vspace{-2mm}
\subsection{Deep Homography Pipelines}
\label{sec:relatedWorks_DeepLearning}
Deep learning pipelines of estimating homography between two images often include: (1) extracting feature map using CNN backbones; (2) predicting one kind of parameterization of homography; (3) computing homography. 
The first deep homography network~\cite{arXiv16} predicts the offsets of four corresponding corners in target image, which comes from a traditional $4$-point parameterization method for
homography computation~\cite{Kanade_TechnicalReport06}.
However, the traditional NDLT-SVD or GPT-LU is also needed to compute $4$ -point homography after network inference.
The subsequent two works~\cite{ICCVW17, ACCV18} follow this pipeline and still require traditional $4$-point homography solvers as a post-processing module.
The first end-to-end deep homography method UDHN~\cite{UDHN_RAL18} integrates TensorDLT into neural network and warp image with the computed homography to measure
photometric loss between two images. The proposed TensorDLT is actually same to the GPU-LU method to solve the formed inhomogeneous linear system $\mathbf{A}_{8*8}\!*\!\mathbf{h}_{8*1}\!\!=\!\!\mathbf{b}_{8*1}$. The only difference is that TensorDLT is differential and implemented under deep learning framework, such as TensorFlow or PyTorch. Most of recent works~\cite{DHDS_CVPR20, CAUDHN_ECCV20, UDIS_TIP21, LocalTrans_ICCV21, DAMG_TCSVT22, IDHN_CVPR22} follow this methodology of predicting the offsets of four corners and adopt TensorDLT as the $4$-point homography solver in their pipelines. 
The main contributions of these methods are reflected in other aspects.
For example, a multi-scale neural network is proposed to predict motion mask and handle dynamic scenes~\cite{DHDS_CVPR20}. 
An unsupervised deep image stitching method is proposed to predict coarse homography for large-baseline scenes and reconstruct the stitched images from feature to pixel~\cite{UDIS_TIP21}.
Another recent work iteratively estimates $4$-point offsets and achieves real-time inference~\cite{IDHN_CVPR22}.
The rightmost column in Table~\ref{tab:relatedWork_pointsHomography} lists the end-to-end deep homography pipelines adopting TensorDLT. Due the concise representation of involving parameter and the form of stratified transformations, the proposed SKS or ACA algorithm is capable to be tensorized in deep homography pipeline to accelerate homography computation.

It is worth noting that there also exist two other kinds of homography parameterization in deep learning pipelines.
The first is inverse compositional Lucas-Kanade (IC-LK)~\cite{LucasKanade_IJCV04}. The works~\cite{CLKN_CVPR17,DLKH_CVPR21} fuse IC-LK into their deep homography pipelines and predict eight parameters of a transformed homography.    
The second is to estimate the coefficients of eight displacement field bases consisting of projective distortion for small baseline scenes~\cite{MotionBasis_ICCV21,UHECAG_CVPR22}.
Although these methods are not directly related to $4$-point homography, the novel parameterization forms which are their emphases are also investigated in this paper.

\vspace{-2mm}
\subsection{Geometry based Homography Decomposition}
\label{sec:relatedWorks_Decom}
In this subsection, we will introduce the evolution of the underlying idea of this paper step by step, based on three works of other researchers and our two previous works. These five works, together with this paper, gradually extend a geometry-based homography decomposition way to camera calibration with various 2D conic patterns and fast homography computation with $4$ points. Different from all previous $4$-point homography methods mentioned in Sec.~\ref{sec:relatedWorks_LinearSystem}, geometry based on methods do not construct the linear system, but utilize homography decomposition to solve the constraints provided by part of corresponding primitives.

Geometry based homography decomposition originally arises from the study of the circular points $\{I,J\}$~\cite{Projective1952}~[p.~33] and their dual conic $[IJ]$. Their projection points denoted by $\{\tilde{I},\tilde{J}\}$ on target plane only have $4$ DOF, but can be used to compute the homography up to a similarity transformation by applying SVD~\cite{Hartley2003Multiple}~[p.~56],
\begin{equation}
   [\tilde{{I}}\tilde{{J}}]=\mathbf{U}[{I}{J}]\mathbf{U}^\top,
    \label{equ:pre_3}
\end{equation}
where $[\tilde{I}\tilde{J}]$ denotes the dual conic of $\{\tilde{I},\tilde{J}\}$ and is a $3$$*$$3$ symmetric matrix with $4$ DOF; $\mathbf{U}\!=\!\mathbf{H}$ up to a similarity transformation.
However, this method is seldom used in practice as the images of two circular points are difficult to be extracted unless the images of two circles~\cite{Wu_IVC06} or the images of one circle and the infinity line can be obtained.
The second row of Table~\ref{tab:relatedWork_decomMethods} shows the characteristics of this method.

The work~\cite{Gurdjos_CVPR06} summarized in the third row of Table~\ref{tab:relatedWork_decomMethods} improves the above method, which utilizes the generalized eigenvalue decomposition (GED) of two confocal conics. Confocal conics is a special kind of pencil of conics~\cite{Projective1952}~[p.~166-170]. 
Denote two confocal conics by $\mathbf{C}_{1}, \mathbf{C}_{2}$. The conic dual to the image of the circular points $[\tilde{I}\tilde{J}]$ can be obtained by utilizing the image of the confocal conics $\tilde{\mathbf{C}}_{1}, \tilde{\mathbf{C}}_{2}$ and their minimum generalized eigenvalue $\lambda_{min}$.
However, this work extracting $[\tilde{I}\tilde{J}]$ from another special 2D pattern still does not extend homography computation to a pair of general imaginary points. 

In one of our previous works~\cite{Mine_MVA12} summarized in the fourth row of Table~\ref{tab:relatedWork_decomMethods}, a transformation mapping a pair of complex conjugate points on the source plane or the target plane to the circular points is constructed under different conics cases. Consequently, a pair of conics intersecting at the pair of complex conjugate points can be transformed to confocal conics or two circles under this projective transformation, which is similar to the above GED method~\cite{Gurdjos_CVPR06}. The difference is that we consider the general conics case and GED is used twice. Thus, this method is called the extended generalized eigenvalue decomposition (EGED). The two intersection points (a pair of complex conjugate points) are denoted by $\{M,N\}$ on the source plane and $\{\tilde{M},\tilde{N}\}$ on the target plane respectively, both of which can be transformed to the circular points $\{I,J\}$ by using SVD. Thus, the homography between a pair of conics and their images can be decomposed into three parts: first projective transformation $\mathbf{H}_{Q}$, middle similarity transformation $\mathbf{H}_{S}$ maintaining the invariance of the circular points, last projective transformation $\mathbf{H}_{P}$. However, this work decomposing homography of general conics pattern is still limited to the imaginary points.

A two real points based homography decomposition (TPHD) method is proposed in another of our previous works~\cite{Guo2017A}, which decomposes a homography into three parts: first projective transformation $\mathbf{H}_{Q}$, middle hyperbolic similarity transformation $\mathbf{H}^{'}_{S}$, last projective transformation $\mathbf{H}_{P}$. See the fifth row of Table~\ref{tab:relatedWork_decomMethods}. The two real intersection points $\{M,N\}$ on the source plane and $\{\tilde{M},\tilde{N}\}$ on the target plane can be transformed to the rectangular hyperbolic
points $\{{I}^{'},{J}^{'}\}$ with the coordinates $[\mp1,1,0]^\top$ under the projective transformation $\mathbf{H}_{Q}$ and $\mathbf{H}_{P}$ by using SVD, respectively. In that work, the problem of camera calibration with two open configurations is handled, which are a conic with a coplanar line and four 3D points. Although the proposed homography decomposition has utilized two real points, the computing efficiency is still not explored. For example, SVD with the computational complexity ${\mathrm{O}}(3^3)$ has to be run twice. 
Moreover, the geometric interpretation of the derived hyperbolic similarity transformation $\mathbf{H}^{'}_{S}$ with four parameters is still unclear, which further limits its practical application.

Another well-known and meaningful homography decomposition method SAP~\cite{liebowitz_CVPR98},~\cite{Hartley2003Multiple}~[p.~42-43] is shown in the sixth row. 
A homography $\mathbf{H}$ is expressed by
\begin{equation}
\begin{split}
\mathbf{H} &= \begin{bmatrix} \mathbf{A} & \mathbf{t} \\
\mathbf{v}^{\mathrm{T}} & v \end{bmatrix} = \mathbf{E}_{S}\mathbf{E}_{A}\mathbf{E}_{P} \\
&= \begin{bmatrix} s\mathbf{R} & \mathbf{t} \\
\mathbf{0}^{\mathrm{T}} & 1 \end{bmatrix}
\begin{bmatrix} \mathbf{K} & \mathbf{0} \\
\mathbf{0}^{\mathrm{T}} & 1 \end{bmatrix}
\begin{bmatrix} \mathbf{I} & \mathbf{0} \\
\mathbf{v}^{\mathrm{T}} & v \end{bmatrix},
\end{split}
\label{equ:SAP}
\end{equation}
where $s$, $\mathbf{R}$ and $\mathbf{t}$ denotes the scale, rotation and translation respectively in the called essence similarity transformation $\mathbf{E}_{S}$; $\mathbf{K}$ denoting the affine components is an upper-triangular matrix whose determinant is $1$ in the essence affine transformation $\mathbf{E}_{A}$; $\mathbf{v}^{\mathrm{T}}$ and $v$ denote the projective components in the essence projective transformation $\mathbf{E}_{P}$; $\mathbf{A}=s\mathbf{R}\mathbf{K}$ denotes affine component.
Since SAP is not based on anchor points, its anchor points related FLOPs does not exist and is categorized into one individual category of primitives configuration in Sec.~\ref{sec:intro}. Moreover, SAP is only suitable for a few of limited scenarios, requiring that the image of the infinity line is firstly available to recover its affine properties, followed by two sets of perpendicular lines to recover the metric properties.

Except the above previous methods, Table~\ref{tab:relatedWork_decomMethods} also lists the pattern configuration, the chosen anchor points (AP), the decomposition formula, and AP-related computational amount (FLOPs) of our SKS and ACA. It is clear that SKS utilizes two similarity transformations $\mathbf{H}_{S_1}$ and $\mathbf{H}_{S_2}$ to manage $\{M,N\}$ and $\{\tilde{M},\tilde{N}\}$, respectively. Thus, SKS avoids the use of SVD and promotes the computational efficiency significantly. 
Furthermore, the proposed ACA method based on three anchor points computes the first and last sub-transformations with less FLOPs.

\vspace{-2mm}
\subsection{Computational Amount and Runtime}
\label{sec:relatedWorks_FLOPs}
Computational amount of an algorithm is typically evaluated by the number of floating-point operations (FLOPs). Compared to the rough statistics of FLOPs (all four arithmetic operations are considered as $1$ FLOPs) provided in the textbooks~\cite{Golub2013, LinearAlgebra}, we follow the Livermore Loops Benchmark~\cite{Livermore86} to count FLOPs. Addition, subtraction and multiplication are considered as $1$ FLOPs, while division is considered as $4$ FLOPs.

In the mathematical society, the computational amount of an algorithm may be evaluated more deeply. 
Take matrix multiplication for an instance. 
Strassen presents that the usual method of computing product of two square matrices (requiring approximately $2n^3$ arithmetical operations) is not optimal~\cite{Strassen1969}. Recently, AlphaTensor~\cite{AlphaTensor2022} is proposed to discover faster matrix multiplication algorithms utilizing reinforcement learning. In Strassen's method and AlphaTensor, for small-size matrices, the number of scalar multiplications decreases, but addition and subtraction operations actually increase.
Therefore, it makes sense for an algorithm to avoid operations with high complexity, whether in theory or in practice.
In this paper, of the four arithmetic operations, division is the most complicated and should be avoided as much as possible.

Besides FLOPs, the runtime of an algorithm on a single core is also influenced by many other hardware factors, such as data transfer and compiler optimization~\cite{Computer_Organization}.
In simple terms, for a program running on a single core of a central processing unit (CPU), compiler optimizations significantly affect the runtime due to the serial scheme of data access and transfer~\cite{microsoftCompiler,IntelCompiler}. Conversely, for programs running on graphics processing units (GPUs), parallel data transfer and processing are more important than other factors.
For a fair comparison in this paper, the homography computation algorithms are tested on CPU with different compiler optimization options and re-implemented on GPU under the same conditions.

\vspace{-2mm}
\section{Derivation of SKS Decomposition}
\label{sec:4Points}
\vspace{-1mm}
This section presents the similarity-kernel-similarity (SKS) decomposition for $4$-point homography computation. Specifically, the overview of decomposition chain is firstly introduced in Sec.~\ref{sec:4pts_overview}. Then, the detailed derivation of the first and last similarity transformations is given in Sec.~\ref{sec:4pts_similartity}, followed by the derivation of the middle kernel transformation shown in Sec.~\ref{sec:4pts_kernel}. 
The computational amount (FLOPs) of SKS with four points is investigated in Sec.~\ref{sec:4pts_FLOPs}.
The $n$-point ($n$$>$$4$) homography computation problem is discussed in Sec.~\ref{sec:SKS_nPoints}.
\begin{figure*}[tb!]
\begin{center}
\includegraphics[width=0.99\textwidth]{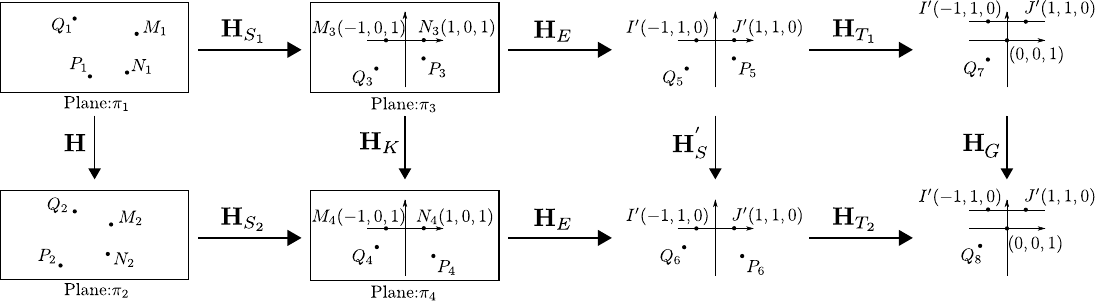}
\vspace{-2mm}
\caption{Sub-transformations in SKS decomposition.}
\label{fig:4Pts_trans}
\end{center}
\vspace{-4mm}
\end{figure*}

\vspace{-2mm}
\subsection{Overview of Decomposition Chain}
\label{sec:4pts_overview}
The first step of SKS is to utilize two points, such as $\{\!M_1,\!N_1\!\}$ and $\{\!M_2,\!N_2\!\}$, to compute a unique similarity transformation on their own plane respectively. See the left part of Fig.~\ref{fig:4Pts_trans}. The similarity transformation deduced by these two points on $\pi_1$ and $\pi_2$ are denoted by $\mathbf{H}_{S_1}$ and $\mathbf{H}_{S_2}$, respectively. 
Under the similarity transformation, the two points will be transformed to two special points with the homogeneous coordinates $[\mp1,0,1]^\top$ (we default that M and N are in the negative and positive directions of the $x$-axis respectively). 
This can be done as both the 2D similarity transformation and two points have $4$ DOF.
The chosen two points are called two anchor points (TAP), because they play a role on normalizing all planar points to a new rectangular plane coordinate system.
The two special points $[\mp1,0,1]^\top$ are thus called canonical TAP.
The remaining unknown component in the homography is called kernel transformation $\mathbf{H}_K$. 
Under $\mathbf{H}_K$, the canonical TAP $[\mp1,0,1]^\top$ should be invariant. 
Moreover, the other two pairs of corresponding points can be utilized to compute $\mathbf{H}_K$ which also has 4 DOF.
The detailed computation process of $\mathbf{H}_K$, including further decomposing it to other transformations (the right part of Fig.~\ref{fig:4Pts_trans}), will be explained in Sec.~\ref{sec:4pts_kernel}.

\subsection{Similarity Transformation based on Two Anchor Points}
\label{sec:4pts_similartity}
Take the two 2D points $\{M_1,N_1\}$ to calculate the similarity transformation $\mathbf{H}_{S_1}$ as an example. The other similarity transformation $\mathbf{H}_{S_2}$ can be similarly calculated from $\{M_2,N_2\}$.

The midpoint $O_1$ of the line segment $M_1N_1$ is given by
\begin{equation}
\begin{split}
& O_1.x = 0.5*(M_1.x+N_1.x),
\\ & 
O_1.y=0.5*(M_1.y+N_1.y),
\end{split}
\label{equ:3_2_1}
\end{equation}

Then the vector $\overrightarrow{O_1N_1}$ is calculated by
\begin{equation}
\begin{split}
&\overrightarrow{O_1N_1}.x=N_1.x-O_1.x, \\ 
&\overrightarrow{O_1N_1}.y=N_1.y-O_1.y. 
\end{split}
\label{equ:3_2_2}
\end{equation}

Instead of directly computing the scale parameter $s_{S_1}$ and the rotation parameter $\theta_{S_1}$ in $\mathbf{H}_{S_1}$, $\mathbf{H}_{S_1}$  can be expressed more easily without the radical and trigonometric operations
\begin{equation}
\mathbf{H}_{S_1} \equiv  \begin{bmatrix}
\overrightarrow{O_1N_1}.x & \overrightarrow{O_1N_1}.y &  \\
 -\overrightarrow{O_1N_1}.y & \overrightarrow{O_1N_1}.x &  \\
   &  & f_{S_1}  \end{bmatrix}
 \begin{bmatrix}
1 &  & -O_1.x \\
  & 1 & -O_1.y \\
   &  & 1  \end{bmatrix},
   \label{equ:3_2_4}
\end{equation}
with
\begin{equation}
f_{S_1} = (\overrightarrow{O_1N_1}.x)^2+(\overrightarrow{O_1N_1}.y)^2,
   \label{equ:fOfS1}
\end{equation}
and we also have
\begin{equation}
\mathbf{H}_{S_1}^{-1} \equiv  \left[\begin{array} {ccc}
\overrightarrow{O_1N_1}.x & -\overrightarrow{O_1N_1}.y & O_1.x \\
 \overrightarrow{O_1N_1}.y & \overrightarrow{O_1N_1}.x & O_1.y \\
   &  & 1  \end{array} \right].
   \label{equ:inverseSimilartiyOfS1}
\end{equation}

Under $\mathbf{H}_{S_1}$, all the 2D points on the plane $\pi_1$ are normalized according to the translation, the scale and the rotation transformations defined by $\{M_1,N_1\}$. $\{M_1,N_1\}$ as the TAP on $\pi_1$ will be transformed to $\{M_3,N_3\}$ (with the coordinates $[\mp1,0,1]^\top$) as the canonical TAP on $\pi_3$. Similarly, $\mathbf{H}_{S_2}$ in Fig.~\ref{fig:4Pts_trans} can also be calculated based on the TAP $\{M_2,N_2\}$ on the plane $\pi_2$. The remaining unknown component $\mathbf{H}_K$ in $\mathbf{H}$ will be introduced in next subsection.

\subsection{Kernel Transformation}
\label{sec:4pts_kernel}
For $\mathbf{H}_K$, a unique property is the invariance of the canonical TAP $[\mp1,0,1]^\top$. We give the following proposition to obtain the expression of $\mathbf{H}_K$.

\vspace{1mm}
\noindent \textbf{Proposition 1}. \textit{The canonical TAP $[\mp1,0,1]^\top$ are fixed under the 2D projective transformation $\mathbf{H}_K$, if and only if $\mathbf{H}_K$ is expressed by}
\begin{equation}
\mathbf{H}_K \equiv
\begin{bmatrix}
a_K & u_K & b_K \\
     & 1 &   \\
 b_K & v_K & a_K \end{bmatrix}.
\label{equ:H_K}
\end{equation}
The proof is shown in Appendix~\ref{appendix:A}. 
\vspace{1mm}

Let the transformed $P_1$ and $Q_1$ under $\mathbf{H}_{S_1}$ on $\pi_3$ be $P_3$ and $Q_3$ respectively. And the transformed $P_2$ and $Q_2$ under $\mathbf{H}_{S_2}$ on $\pi_4$ are denoted by $P_4$ and $Q_4$ respectively. $\mathbf{H}_K$ with 4 DOF can be calculated by these two pairs of corresponding points $\{P_3{\stackrel{\mathbf{H}_K}{\longrightarrow}}P_4\}$ and $\{Q_3{\stackrel{\mathbf{H}_K}{\longrightarrow}}Q_4\}$. 
$\mathbf{H}_K$ is further associated with the hyperbolic similarity transformation~\cite{Guo2017A} as follows.

We notice that the points $[\mp1,0,1]^\top$ can be transformed to the rectangular hyperbolic points $\{{I}^{'}\!,{J}^{'}\}$ ($=[\mp1,1,0]^\top$) under an elementary transformation $\mathbf{H}_{E}$, which is expressed by
\begin{equation}
\mathbf{H}_{E} =
\begin{bmatrix}
1 &  &  \\
   & & 1  \\
  &  1 & \end{bmatrix}.
\end{equation}
\begin{table*}[tb!]
\begin{center}
\caption{FLOPs of each step in the proposed SKS method with $4$ points. $*$ denotes the operation of matrix multiplication. After obtaining the homography up to a scale, normalization can be done with extra $12$ FLOPs at last.
Detailed FLOPs notations can be seen in the released source code.}
\label{tab:FLOPs_ours_4pts} 
\vspace{3mm}
\resizebox{0.99\textwidth}{!}{
\begin{tabular}{|l||c|c|c|c|c|c|c|c|c|c|c|c|c|c|c|c|}
\hline
\multirow{3}{*}{Step}  & \multicolumn{2}{c|}{$\mathbf{H}_{S_1}$}  & \multicolumn{2}{c|}{$\mathbf{H}_{S_2}$}  & \multicolumn{2}{c|}{$\mathbf{H}_{T_1}$}  &  \multicolumn{2}{c|}{$\mathbf{H}_{T_2}$}  & \multicolumn{4}{c|}{$\mathbf{H}_{K}$}  & \multicolumn{2}{c|}{$*$} & \multirow{3}{*}{Sum} &
\multirow{3}{*}{w. Norm.} \\
\cline{2-15}
 & \begin{scriptsize}$\overrightarrow{O_1N_1}$\end{scriptsize}  & \tabincell{c}{$f_{S_1}$} & \begin{scriptsize}$\overrightarrow{O_2N_2}$\end{scriptsize}  & \tabincell{c}{$f_{S_2}$} & $P_3$ & $P_5$ & $P_4$ & $P_6$ & $Q_7$ &  $Q_8$ & \tabincell{c}{$a_{K}$ \\ $b_{K}$} & \tabincell{c}{$u_{K}$ \\ $v_{K}$} & \begin{scriptsize} $\mathbf{H}^{-1}_{S_2}\!*\!\mathbf{H}_{K}$ \end{scriptsize} & \begin{scriptsize}$*\mathbf{H}_{S_1}$ \end{scriptsize}  & & \\
\hline
SKS (complete)   & 6 & 3 & 6 & 3 & 8 & 6 & 8 & 6 & 14 & 14 & 16 & 8 & 20 & 39 & 157 & 169\\
\hline
SK (simplified)  & 0 & 0 & 6 & 3 & 0 & 0 & 8 & 6 & 0 & 14 & 8 & 8 & 20 & 0 & 73 & 85\\
\hline
\end{tabular}}
\end{center}
\vspace{-5mm}
\end{table*}

Then, the hyperbolic similarity transformation $\mathbf{H}^{'}_{S}$, under which $\{{I}^{'}\!,{J}^{'}\}$ are fixed, is further decomposed as follows for computing convenience.
\begin{align}\label{equ:3}\notag
\mathbf{H}^{'}_{S} &=\!
\mathbf{H}^{-1}_{T_2}\mathbf{H}_{G}\mathbf{H}_{T_1}\\
&=\!  
\begin{bmatrix}
1 & & P_6.x  \\
 & 1 & P_6.y  \\ 
 & & 1 \end{bmatrix}\!
 \begin{bmatrix}
a_G & b_G &   \\
 b_G & a_G &  \\
 & & 1 \end{bmatrix}\!
\begin{bmatrix}
1 & & -P_5.x  \\
 & 1 & -P_5.y \\
 & & 1 \end{bmatrix},
\end{align}
\noindent where $P_5$ and $P_6$ are the transformed points of $P_3$ and $P_4 $ under $\mathbf{H}_{E}$, respectively; $\mathbf{H}_{T_1}$ and $\mathbf{H}_{T_1}$ denote the 2D translation transformations; $\mathbf{H}_{G}$ denoting hyperbolic rotation and scale can be computed by solving the following linear equations in two variables:
\begin{equation}
\begin{split}
& Q_8.x = a_G*Q_7.x + b_G*Q_7.y, \\
& Q_8.y = b_G*Q_7.x + a_G*Q_7.y,
\end{split}
\label{equ:4}
\end{equation}
\noindent where $Q_7$ is the transformed point of $Q_3$ under $\mathbf{H}_{T_1}\!*\!\mathbf{H}_{E}$, and $Q_8$ is the transformed point of $Q_4$ under $\mathbf{H}_{T_2}\!*\!\mathbf{H}_{E}$. 

Fig.~\ref{fig:4Pts_trans} depicts each sub-transformation of the proposed SKS decomposition. Obviously, $\mathbf{H}_{S_1}$ and $\mathbf{H}_{S_2}$ do not influence the shape of the quadrangles on two planes. Therefore the transformation $\mathbf{H}_{K} = \mathbf{H}_{E}^{-1}\mathbf{H}^{'}_{S}\mathbf{H}_{E}$ results in the real projective distortion. We call $\mathbf{H}_K$ \textit{the kernel transformation}, in which four variables are calculated by
\begin{equation}
\begin{split}
a_K &= a_G, \quad b_K = b_G,     \\
u_K &= P_6.x-a_K\!*\!P_5.x-b_K\!*\!P_5.y, \\
v_K &= P_6.y-b_K\!*\!P_5.x-a_K\!*\!P_5.y.
\end{split}
\end{equation}

Finally, the homography $\mathbf{H}$ is expressed by the above \textit{similarity-kernel-similarity} (SKS) decomposition form:
\begin{equation}
\mathbf{H}=\mathbf{H}_{S_2}^{-1}\mathbf{H}_K\mathbf{H}_{S_1}=\mathbf{H}_{S_2}^{-1}\mathbf{H}_{E}^{-1}\mathbf{H}^{'}_{S}\mathbf{H}_{E}\mathbf{H}_{S_1}.
\label{equ:HS}
\end{equation}

The distinction between SKS (Eq.~\ref{equ:HS}) and TPHD~\cite{Guo2017A} is that SKS uses a simpler and more efficient expression $\mathbf{H}_{S_1}\mathbf{H}_{E}$ to transform the two anchor points to the rectangular hyperbolic points $\{{I}^{'}\!,{J}^{'}\}$. While in TPHD, it needs to repeat the calculation of SVD for a $3$$*$$3$ matrix (a projective transformation) twice, which is not ignorable compared with a small number of arithmetic calculations of SKS.
Moreover, the kernel transformation $\mathbf{H}_K$ in Eq.~\ref{equ:HS} denoting the projective distortion has more algebraic and geometrical properties than $\mathbf{H}^{'}_{S}$.

\vspace{-2mm}
\subsection{FLOPs Analysis of SKS}
\label{sec:4pts_FLOPs}
In this subsection, we analyze floating-point operations (FLOPs) of the proposed SKS method under $4$ points pattern. The counting method of FLOPs has been introduced in Sec.~\ref{sec:relatedWorks_FLOPs}. Owing to our simple expression and calculation of a homography in Eq.~\ref{equ:HS}, only a small number of FLOPs is required for each step in SKS.
The specific FLOPs statistics of the proposed SKS method with $4$ points is further divided into two situations according to different practical applications.

\textrm{i}) \textit{The homography between two images $\mathcal{I}_1$ and $\mathcal{I}_2$}, i.e., $\{\mathbf{H}\!\!:\mathcal{I}_1\!\!\to\!\!\mathcal{I}_2\}$.
This situation often occurs in image stitching or visual simultaneous localization and mapping (SLAM), where a sequence of images is captured by a moving camera.
In this situation, each sub-transformation of the SKS shown in Fig.~\ref{fig:4Pts_trans} should be calculated.
We call this situation the complete SKS.
FLOPs of each manipulation step in the complete SKS is shown in Table~\ref{tab:FLOPs_ours_4pts}. 
For example, in the complete SKS, the computation of $\mathbf{H}^{'}_{S}$ involves four sub-steps, which are computing $Q_7$, computing $Q_8$, solving \{$a_K,\!b_K$\} and computing the multiplication of the three sub-transformations. The sum FLOPs of these steps in SKS is $157$. If normalization of homography is required (generally based on the lower right element $h_{33}$ of $\mathbf{H}$), there are $1$ additional division and $8$ multiplication operations ($12$ FLOPs).

\textrm{ii}) \textit{The homography between an object plane $\mathcal{O}$ and one its image $\mathcal{I}$}, i.e., $\{\mathbf{H}\!\!:\!\mathcal{O}\!\!\to\!\!\mathcal{I}\}$. This situation often occurs in camera calibration and metric rectification with a known planar pattern. In this situation, the pattern (e.g., chessboard or QR code) with known geometry is captured to obtain the homography between the object plane and one its image. 
Changing the object coordinate system into a new one determined by two pre-selected anchor points in advance, we no longer need to calculate the similarity transformation $\mathbf{H}_{S_1}$. 
The homography up to a similarity can be utilized for the computation of intrinsic parameters in camera calibration, or recover planar metric information~\cite{Hartley2003Multiple}~[p.~57].
This situation is thus called the simplified SK. Most applications under this situation (e.g., camera calibration and object detection) do not care about the change of the origin and rotation of the object coordinate system. When the real translation in the relative pose needs to be calculated, we only need to multiply the final result by a pre-known scale factor.
Under this situation, the homography $\bar{\mathbf{H}}$ to be solved satisfies
\begin{equation}
\bar{\mathbf{H}} = \mathbf{H}\mathbf{H}_{S_1}^{-1} = \mathbf{H}_{S_2}^{-1}\mathbf{H}_K.
\label{equ:SK}
\end{equation}

In addition, other calculation steps are also simplified, such as the calculation of $\mathbf{H}_{T_1}$ and the operation of matrix multiplication.
The FLOPs of each step in this simplified SK decomposition is also listed in Table~\ref{tab:FLOPs_ours_4pts}.
Compared to the complete SKS, the simplified SK reduces the calculation amount by about half.

FLOPs of other $4$-point homography computation methods, including NDLT-SVD~\cite{Hartley2003Multiple}, HO-SVD~\cite{HO_BMVC05},  GPT-LU~\cite{OpenCV_GPT} and RHO-GE~\cite{Bazargani2015Fast}, have been given in Table~\ref{tab:relatedWork_pointsHomography}. 
Compared to NDLT-SVD, HO-SVD, GPT-LU and RHO-GE, SKS represents effective speedup about $162$\textit{x}, $11$\textit{x}, $12$\textit{x} and $1.3$\textit{x} in terms of FLOPs respectively.
These comparison of FLOPs are basically consistent with the experimental results of rumtime on CPU and GPU shown in Sec.~\ref{sec:exp_CPU} and Sec.~\ref{sec:exp_GPU} respectively.

\vspace{-2mm}
\subsection{Discussion of $n$-Point Homography Computation}
\label{sec:SKS_nPoints}
In this subsection, we briefly discuss the problem of $n$-Point homography computation. 
NDLT-SVD~\cite{Hartley2003Multiple}~[p.~109] and HO-SVD~\cite{HO_BMVC05} can handle $n$-point configuration, but at the cost of a lot of redundancy in their $4$-point homography computation.
Although SKS can be adjusted to process $n$-point pattern to bring speed improvements, the stratified computation strategy is unchanged. Thus, each sub-transformation in SKS computed with a part of $n$ point correspondences is sub-optimal. As a result, for general situations where speed is not the first factor, we still recommend using the HO-SVD method to compute $n$-point homography.
As pointed out in Sec.~\ref{sec:relatedWorks_RANSAC}, $n$-point homography computation is not needed in the standard RANSAC~\cite{Hartley2003Multiple}~[p.~123].
Although there exists the method~\cite{chum2003locally} using $n$-point homography computation to replace final global optimization algorithms, the computational cost of the RANSAC framework mainly depends on the calculation and verification of $4$-point homography, especially facing high outlier ratios.

\vspace{-2mm}
\section{Affine-Core-Affine (ACA) Decomposition}
\label{sec:ACA}
This section presents an improved affine-core-affine (ACA) decomposition for $4$-point homography computation. 
Specifically, other choices of canonical two anchor points is introduced in Sec.~\ref{sec:ACA_otherTAP}. Then, the detailed derivation of the first and last affine transformations based on three anchor points is given in Sec.~\ref{sec:ACA_affine}, followed by the derivation of the middle core transformation shown in Sec.~\ref{sec:ACA_core}. 
The computation amount (FLOPs) of ACA with $4$ points is demonstrated in Sec.~\ref{sec:ACA_FLOPs}. 
Further polynomial expression of each element of homography in given in Sec.~\ref{sec:ACA_polynomial}. 

\vspace{-3mm}
\subsection{Other Choices of Canonical Two Anchor Points}
\label{sec:ACA_otherTAP}
In the derivation of the SKS decomposition, we set $[\mp1,0,1]^\top$ to be the canonical TAP to compute two similarity transformations on the source and target planes.
However, the canonical TAP can also be selected at other positions. Consequently, the geometric meaning and algebraic calculation of the new homography decomposition will be a little different, especially for the kernel transformation $\mathbf{H}_K$. Denote the similarity transformation mapping $[\mp1,0,1]^\top$ to the new TAP by $\mathbf{H}_{S_3}$. The SKS decomposition based on new canonical TAP is expressed by
\begin{equation}
\mathbf{H} = \underbrace{\mathbf{H}_{S_2}^{-1}\mathbf{H}_{S_3}^{-1}}_{\text{new}\:\mathbf{H}_{S_2}^{-1}} * \underbrace{\mathbf{H}_{S_3}\mathbf{H}_K\mathbf{H}_{S_3}^{-1}}_{\text{new}\:\mathbf{H}_{K}} * \underbrace{\mathbf{H}_{S_3}\mathbf{H}_{S_1}}_{\text{new}\:\mathbf{H}_{S_1}}.
\label{equ:SKS_newTAP}
\end{equation}

For example, if the canonical TAP are set to be $[0,\mp1,1]^\top$, most of sub-transformations in the decomposition chain shown in Fig.~\ref{fig:4Pts_trans} remain the same except $\mathbf{H}_{E}$ and $\mathbf{H}_{K}$. $\mathbf{H}_{E}$ will play the role of transforming $[0,\mp1,1]^\top$ to the rectangular hyperbolic points $\{{I}^{'}\!,{J}^{'}\}$. The new $\mathbf{H}_{K}$ will be expressed by 
\begin{equation}
\begin{split}
\mathbf{H}_{K} &\equiv \begin{bmatrix}
 & -1 &  \\
1 &  &  \\
   &  & 1  \end{bmatrix}\begin{bmatrix}
a_K & u_K & b_K  \\
 & 1 &  \\
  b_K & v_K & a_K  \end{bmatrix}\begin{bmatrix}
 & 1 &  \\
-1 &  &  \\
   &  & 1  \end{bmatrix} \\
   &\equiv \begin{bmatrix}
1 &  &   \\
-u_K & a_K & b_K \\
  -v_K & b_K & a_K  \end{bmatrix}.
   \end{split}
   \label{equ:new1_H_K}
\end{equation}

Another mapping choice is to set the canonical TAP be $\{[0,0,1]^\top,[1,0,1]^\top\}$. Therefore, $\mathbf{H}_{S_3}$ is given by
\begin{equation}
\begin{split}
\mathbf{H}_{S_3} &\equiv \begin{bmatrix}
0.5 & 0.5 &  \\
 & 0.5 &  \\
   &  & 1  \end{bmatrix},
   \end{split}
   \label{equ:new2_H_S3}
\end{equation}
and the new $\mathbf{H}_{K}$ is given by
\begin{equation}
\begin{split}
\mathbf{H}_{K} &\equiv \begin{bmatrix}
a_K\!+\!b_K & u_K\!+\!v_K &  \\
 & 1 &  \\
  2\!*\!b_K & 2\!*\!v_K & a_K\!-\!b_K  \end{bmatrix} \\
  &\equiv \begin{bmatrix}
1 & u_K^{'} &  \\
 & v_K^{'} &  \\
  1\!-\!a_K^{'} & b_K^{'} & a_K^{'}  \end{bmatrix},
   \end{split}
   \label{equ:new2_H_K}
\end{equation}
where the four parameters are re-named and re-scaled as follows
\begin{equation}
\begin{split}
a_K^{'} = \frac{a_K-b_K}{a_K+b_K}, \quad  &b_K^{'} = \frac{2*b_K}{a_K+b_K}, \\
u_K^{'} = \frac{u_K+v_K}{a_K+b_K}, \quad  &v_K^{'} = \frac{2*v_K}{a_K+b_K}.
   \end{split}
\end{equation}

\vspace{-2mm}
\subsection{Affine Transformation based on Three Anchor Points}
\label{sec:ACA_affine}
\begin{minipage}{0.5\textwidth}
Compared with varying the position of the canonical two anchor points, choosing three anchor points (HAP) to compute the first and last transformations will bring more significant changes. Let the canonical HAP be $\{[0,0,1]^\top,[1,0,1]^\top,[0,1,1]^\top\}$. Three points on the source plane $\pi_1$ and the target plane $\pi_2$, such as $\{\!M_1,\!N_1\!,\!P_1\!\}$ and $\{\!M_2,\!N_2\!,\!P_2\!\}$, can be utilized to compute a unique affine transformation with $6$ DOF on their own plane respectively. Take $\{\!M_1,\!N_1\!,\!P_1\!\}$ to calculate the affine transformation $\mathbf{H}_{A_1}$ on $\pi_1$ as an example. 
It is not difficult to obtain
\end{minipage}
\begin{equation}
\mathbf{H}_{A_1} \!\equiv\! \begin{bmatrix}
\overrightarrow{M_1P_1}.y & -\overrightarrow{M_1P_1}.x &  \\
 -\overrightarrow{M_1N_1}.y & \overrightarrow{M_1N_1}.x &  \\
   &  & f_{A_1}  \end{bmatrix} \!
 \begin{bmatrix}
1 &  & -M_1.x \\
  & 1 & -M_1.y \\
   &  & 1  \end{bmatrix}, 
   \label{equ:AffineExpresion}
\end{equation}
with
\begin{equation}
f_{A_1} = \overrightarrow{M_1N_1}.x*\overrightarrow{M_1P_1}.y-\overrightarrow{M_1N_1}.y*\overrightarrow{M_1P_1}.x,
\label{equ:f_A_1}
\end{equation}
and
\begin{equation}
\mathbf{H}_{A_1}^{-1} \equiv \begin{bmatrix}
\overrightarrow{M_1N_1}.x & \overrightarrow{M_1P_1}.x & M_1.x \\
 \overrightarrow{M_1N_1}.y & \overrightarrow{M_1P_1}.y & M_1.y \\
   &  & 1  \end{bmatrix}.
   \label{equ:inverseAffineOfA1}
\end{equation}

Similarly, the affine transformation $\mathbf{H}_{A_2}$ on $\pi_2$ can also be computed.
\begin{table*}[t]
\begin{center}
\caption{FLOPs of each step in ACA with $4$ points. $*$ denotes the operation of matrix multiplication. Detailed FLOPs notation can be seen in the released source code. Without normalization, all other steps to compute the homography up to a scale do not involve any division operations.}
\label{tab:FLOPs_ACA_4pts} 
\vspace{3mm}
\resizebox{0.99\textwidth}{!}{
\begin{tabular}{|l||c|c|c|c|c|c|c|c|c|c|c|c|c|c|c|c|c|}
\hline
\multirow{2}{*}{Step}  & \multicolumn{3}{c|}{$\mathbf{H}_{A_1}$}  & \multicolumn{3}{c|}{$\mathbf{H}_{A_2}$}  &  \multicolumn{7}{c|}{$\mathbf{H}_{C}$}  & \multicolumn{2}{c|}{$*$} & \multirow{3}{*}{Sum} & \multirow{3}{*}{w. Norm.} \\
\cline{2-16}
 & \begin{scriptsize}$\overrightarrow{M_1N_1}$\end{scriptsize}  & \begin{scriptsize}$\overrightarrow{M_1P_1}$\end{scriptsize} & $f_{A_1}$ & \begin{scriptsize}$\overrightarrow{M_2N_2}$\end{scriptsize}  & \begin{scriptsize}$\overrightarrow{M_2P_2}$\end{scriptsize} & $f_{A_2}$ &  \begin{scriptsize}$\overrightarrow{M_1N_1}$\end{scriptsize} & \begin{scriptsize}$\overrightarrow{M_2N_2}$\end{scriptsize} & $Q_3$ &  $Q_4$ & $t_1$ & $t_2$ & $c_{11},c_{22},c_{33}$ & \begin{scriptsize} $\mathbf{H}^{-1}_{A_2}*\mathbf{H}_{C}$ \end{scriptsize} & \begin{scriptsize}$*\mathbf{H}_{A_1}$ \end{scriptsize}  &  & \\
\hline
ACA   & 2 & 2 & 3 & 2 & 2 & 3 & 2 & 2 & 6 & 6 & 2 & 2 & 8 & 10 & 33 & 85 & 97\\
\hline
\end{tabular}}
\end{center}
\vspace{-5mm}
\end{table*}
\begin{figure}[t]
\begin{center}
\includegraphics[width=0.42\textwidth]{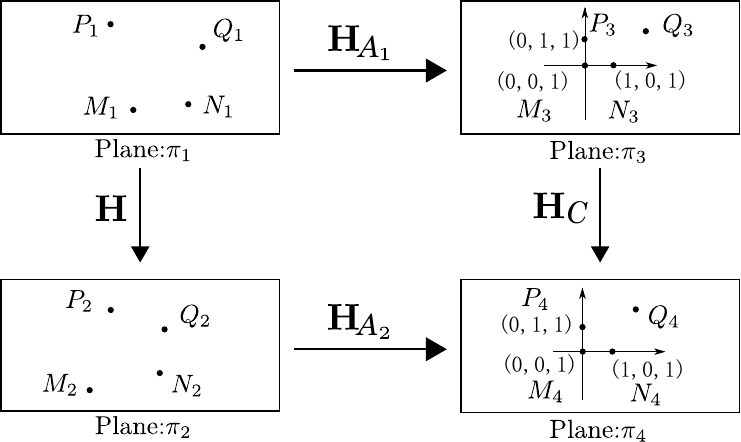}
\vspace{-3mm}
\caption{Sub-transformations in ACA decomposition.}
\label{fig:ACA_trans}
\end{center}
\vspace{-5mm}
\end{figure}

\vspace{-2mm}
\subsection{Core Transformation}
\label{sec:ACA_core}
Similar to SKS, we can obtain a homography decomposition with three components after obtaining the two affine transformations on the source and target planes.
As shown in Fig.~\ref{fig:ACA_trans}, under $\mathbf{H}_{A_1}$ and $\mathbf{H}_{A_2}$, the plane $\pi_1$ and $\pi_2$ are transformed to two new planes $\pi_3$ and $\pi_4$, respectively. The projective transformation between $\pi_3$ and $\pi_4$ is denoted by $\mathbf{H}_C$.
Because of the invariance of the canonical HAP, the following proposition is reached to obtain the expression of $\mathbf{H}_C$.

\vspace{1mm}
\noindent \begin{minipage}{0.5\textwidth}
\textbf{Proposition 4}. \textit{The canonical three anchor points
$\{[0,0,1]^\top, [1,0,1]^\top,[0,1,1]^\top\}$ are fixed under the 2D projective transformation $\mathbf{H}_C$, if and only if $\mathbf{H}_C$ is expressed by}
\end{minipage}

\begin{equation}
\mathbf{H}_C \equiv 
\left[\begin{array} {ccc}
a_C &  &  \\
     & b_C &   \\
 a_C\!-\!1 & b_C\!-\!1 & 1 \end{array}\right].
\label{equ:H_C}
\end{equation}
The proof is similar to it in Appendix~\ref{appendix:A} and omitted. 
\vspace{1mm}

Since $\mathbf{H}_C$ imposes the projective distortion between two affine normalization planes, we call $\mathbf{H}_C$ \textit{the core transformation}. Therefore, a homography $\mathbf{H}$ is expressed by the \textit{affine-core-affine} (ACA) decomposition form:
\begin{equation}
\mathbf{H} = \mathbf{H}_{A_2}^{-1}\mathbf{H}_C\mathbf{H}_{A_1}.
\label{equ:ACA}
\end{equation}

Obviously, $\mathbf{H}_C$ with 2 DOF can be computed from $\{Q_3,Q_4\}$, which are transformed from the fourth point correspondence $\{Q_1,Q_2\}$ under $\mathbf{H}_{A_1}$ and $\mathbf{H}_{A_2}$ respectively. Specifically, the point $Q_3$ denoted by $[Q_3.x,Q_3.y,f_{A_1}]^\top$ is obtained by
\begin{equation}
\begin{split}
\begin{bmatrix} Q_3.x \\
Q_3.y \\ f_{A_1} \end{bmatrix} &=  \mathbf{H}_{A_1}*\begin{bmatrix} Q_1.x \\
Q_1.y \\ 1 \end{bmatrix} \\
&= \begin{bmatrix} \overrightarrow{M_1Q_1}.x\!*\!\overrightarrow{M_1P_1}.y-\overrightarrow{M_1Q_1}.y\!*\!\overrightarrow{M_1P_1}.x \\
\overrightarrow{M_1N_1}.x\!*\!\overrightarrow{M_1Q_1}.y-\overrightarrow{M_1N_1}.y\!*\!\overrightarrow{M_1Q_1}.x \\ f_{A_1} \end{bmatrix},
\end{split}
   \label{equ:Q3}
\end{equation}
where $Q_3.x$ and $Q_3.y$ denotes the coordinates of $Q_3$ up to a scale $f_{A_1}$ for computational convenience. 

Similarly, the point $Q_4$ denoted by $[Q_4.x,Q_4.y,f_{A_2}]^\top$ can be obtained. As $Q_4$ is the projection of $Q_3$ under the core transformation $\mathbf{H}_C$, $a_C$ and $b_C$ in Eq.~\ref{equ:H_C} are solved as:
\begin{equation}
\begin{split}
a_C =  \frac{t_1*Q4.x}{t_2*Q3.x}, \quad
b_C =  \frac{t_1*Q4.y}{t_2*Q3.y}, \quad
\end{split}
   \label{equ:a_C}
\end{equation}
with 
\begin{equation}
\begin{split}
t_1\!=\!f_{A_1}\!-\!Q_3.x\!-\!Q_3.y, \quad t_2\!=\!f_{A_2}\!-\!Q_4.x\!-\!Q_4.y.
\end{split}
   \label{equ:t1}
\end{equation}

To avoid the division operations in the above equations, we change the expression of $\mathbf{H}_C$ up to a scale as follows,
\begin{equation}
\mathbf{H}_C \equiv \begin{bmatrix} c_{11} & & \\
& c_{22} &  \\ c_{11}\!-\!c_{33} & c_{22}\!-\!c_{33} & c_{33} \end{bmatrix},
\label{equ:H_C_2}
\end{equation}
with
\begin{equation}
\begin{split}
c_{11} = t_1*Q_3.y*Q_4.x, \\
c_{22} = t_1*Q_3.x*Q_4.y, \\
c_{33} = t_2*Q_3.x*Q_3.y.
\label{equ:C_11}
\end{split}
\end{equation}

Substituting the above expression of $\mathbf{H}_C$ into Eq.~\ref{equ:ACA}, the homography $\mathbf{H}$ up to a scale is finally computed without any division operations. 

\vspace{-2mm}
\subsection{FLOPs Analysis of ACA}
\label{sec:ACA_FLOPs}
Similar to the computation analysis of SKS shown in Sec.~\ref{sec:4pts_FLOPs}, we give the number of FLOPs in each step of ACA shown in Table~\ref{tab:FLOPs_ACA_4pts}. FLOPs of both $\mathbf{H}_{A_1}$ and $\mathbf{H}_{A_2}$ are $7$ according to Eq.~\ref{equ:AffineExpresion} and Eq.~\ref{equ:f_A_1}. FLOPs of calculating $\mathbf{H}_{C}$ is $28$ according to Eq.~\ref{equ:Q3}, Eq.~\ref{equ:t1} and Eq.~\ref{equ:C_11}. The multiplication operation of these sub-transformations costs $43$ FLOPs. Without consideration of the final normalization (which has $12$ FLOPs), ACA can compute a homography up to a scale with only $85$ FLOPs and do not involve any division operations.
Compared to the previous $4$-point homography methods NDLT-SVD, HO-SVD, GPT-LU and RHO-GE shown in Table~\ref{tab:relatedWork_pointsHomography}, the total $97$ FLOPs of ACA is about $0.35\%$, $5.4\%$, $5.0\%$ and $43\%$ of their FLOPs, respectively.
These theoretical speedups are also reflected by the experimental results running on CPU and GPU shown in Sec.~\ref{sec:experiments}.

Here we only analyze the FLOPs of ACA for the mapping between four general points on source and target planes. When the quadrilateral on the source plane is a square or rectangle (often seen in deep homography pipelines), the calculation process is further simplified, which will be illustrated in Sec.~\ref{sec:app_tensorACA}.

\vspace{-3mm}
\subsection{Polynomial Expression of Homography}
\label{sec:ACA_polynomial}
This subsection describes another algebraic advantage of the proposed ACA decomposition.
Owing to the extremely simple expression of each sub-transformation, we can represent each element of a homography by the input variables ($16$ coordinates provided by four point correspondences) in polynomial form.

See Eq.~\ref{equ:AffineExpresion} and  Eq.~\ref{equ:inverseAffineOfA1}. It is clear that $\mathbf{H}_{A_1}$ and $\mathbf{H}_{A_2}^{-1}$ can be directly expressed with the input $16$ coordinates. Substituting $\mathbf{H}_C$ in Eq.~\ref{equ:H_C} into the ACA decomposition in Eq.~\ref{equ:ACA}, after some manipulations, the column vector form of $\mathbf{H}$ is given by
\vspace{-2mm}

\begin{tiny}
\begin{equation}
\begin{split}
\begin{bmatrix}
h_{11} \\
h_{21} \\
h_{31} \\
h_{12} \\
h_{22} \\
h_{32} \\
h_{13} \\
h_{23} \\
h_{33} \\
\end{bmatrix} \!=\!
\begin{bmatrix}
N_2.x(P_1.y\!-\!M_1.y) & P_2.x(M_1.y\!-\!N_1.y) &
M_2.x(N_1.y\!-\!P_1.y) \\
N_2.y(P_1.y\!-\!M_1.y) & P_2.y(M_1.y\!-\!N_1.y) &
M_2.y(N_1.y\!-\!P_1.y) \\
(P_1.y\!-\!M_1.y) & (M_1.y\!-\!N_1.y) &
(N_1.y\!-\!P_1.y) \\
N_2.x(M_1.x\!-\!P_1.x) & P_2.x(N_1.x\!-\!M_1.x) &
M_2.x(P_1.x\!-\!N_1.x) \\
N_2.y(M_1.x\!-\!P_1.x) & P_2.y(N_1.x\!-\!M_1.x) &
M_2.y(P_1.x\!-\!N_1.x) \\
(M_1.x\!-\!P_1.x) & (N_1.x\!-\!M_1.x) &
(P_1.x\!-\!N_1.x) \\
N_2.x*f_1 & P_2.x*f_2 &
M_2.x*f_{A_1} \\
N_2.y*f_1 & P_2.y*f_2 &
M_2.y*f_{A_1} \\
f_1 & f_2 &
f_{A_1} \\
\end{bmatrix}  \\
*\begin{bmatrix}
c_{11} \\
c_{22}  \\
c_{33} 
\end{bmatrix},
\label{equ:H_11}
\end{split}
\end{equation}
\end{tiny}
\vspace{-2mm}

\noindent where $f_1$ and $f_2$, as well as $f_{A_1}$ in Eq.~\ref{equ:f_A_1}, are second degree polynomials of input coordinates, which are expressed by,
\begin{equation}
\begin{split}
f_1 &= P_1.x*M_1.y-P_1.y*M_1.x,\\
f_2 &= M_1.x*N_1.y-M_1.y*N_1.x.
\label{equ:f_1}
\end{split}
\end{equation}

Meanwhile, it can be seen from Eq.~\ref{equ:Q3} and Eq.~\ref{equ:t1} that $Q_3.x$, $Q_3.y$, $Q_4.x$, $Q_4.y$, $t_1$ and $t_2$ are also second degree polynomials.
Therefore, $c_{11}$, $c_{22}$, $c_{33}$ in Eq.~\ref{equ:C_11} are sixth degree polynomials. Thus it is straightforward that each element in the vector form of a homography is a polynomial, with $8$, $8$, $7$, $8$, $8$, $7$, $9$, $9$, $8$ degree respectively, which is expressed by

\begin{equation}
\mathbf{H} = \begin{bmatrix} \mathcal{F}_{11}^8 & \mathcal{F}_{12}^8 & \mathcal{F}_{13}^9 \\
\mathcal{F}_{21}^8 & \mathcal{F}_{22}^8 & \mathcal{F}_{23}^9  \\ \mathcal{F}_{31}^7 & \mathcal{F}_{32}^7 & \mathcal{F}_{33}^8 \end{bmatrix},
\end{equation}
where $\mathcal{F}^i$ denotes an $i$-th degree polynomial.

To our knowledge, this is the first time to obtain the polynomial expression of a homography directly with the input $16$ coordinates. An interesting observation is that if there is a division operation in the calculation process, it may be impossible to obtain the polynomial expression of a homography.

\vspace{-2mm}
\section{Extension and Applications}
\label{sec:application}
The above two sections describe the proposed SKS and ACA methods respectively, in both of which $4$-point homography is computed with clear geometric meaning and fast calculation. In this section, we will introduce their further extension and applications. Specifically, how SKS and ACA extend the existing SAP decomposition is illustrated in Sec.~\ref{sec:app_linkSAP}.
The SKS and ACA decomposition for a 2D affine transformation is given in Sec.~\ref{sec:app_affineDecomposition}. The tensorized ACA (TensorACA) used in end-to-end deep homography estimation is proposed in Sec.~\ref{sec:app_tensorACA}. 

\subsection{Extending SAP Decomposition}
\label{sec:app_linkSAP}
Consider our SKS decomposition shown in Eq.~\ref{equ:SKS_newTAP} and Eq.~\ref{equ:new2_H_K}. Assuming $\mathbf{H}_{S_1}$ is an identity matrix, one equivalent decomposition is derived by us,
\begin{equation}
\begin{split}
\mathbf{H} &=  \mathbf{H}_{S_2}^{-1}\mathbf{H}_{K} \\
&= \mathbf{H}_{S_2}^{-1}\begin{bmatrix}
1 & u_K^{'} &  \\
 & v_K^{'} &  \\
   &  & 1  \end{bmatrix}
  \begin{bmatrix}
1 &  &  \\
 & 1 &  \\
  1\!-\!a_K^{'} & b_K^{'} & a_K^{'}  \end{bmatrix},
\end{split}
\label{equ:SAP_SKS}
\end{equation}
where it can be seen that SAP is just one special form of the proposed SKS decomposition.

Furthermore, if combining the essence similarity and affine transformations into a general affine transformation, the SAP decomposition will have one fixed form,
\begin{equation}
\begin{split}
\mathbf{H} \!=\!  \begin{bmatrix}
h_{11}\!-\!h_{31}\frac{h_{13}}{h_{33}} & h_{12}-h_{32}\frac{h_{13}}{h_{33}} & \frac{h_{13}}{h_{33}} \\
h_{21}\!-\!h_{31}\frac{h_{23}}{h_{33}} & h_{22}-h_{32}\frac{h_{23}}{h_{33}} & \frac{h_{23}}{h_{33}} \\
   &  & 1  \end{bmatrix} \!\!
  \begin{bmatrix}
1 &  &  \\
 & 1 &  \\
  h_{31} & h_{32} & h_{33}  \end{bmatrix},
\end{split}
\label{equ:AP}
\end{equation}
where this expression is also equivalent to the proposed ACA decomposition, if we assume $\mathbf{H}_{A_1}$ is an identity matrix and make an adequate transformation.

The fixed forms of Eq.~\ref{equ:SAP} and Eq.~\ref{equ:AP} greatly limit the practical use of the SAP decomposition. 
However, \textit{the proposed SKS and ACA decomposition can be regarded as an extension of SAP, enjoying the notable merits of dealing with general primitives and fast calculation}.

\vspace{-2mm}
\subsection{Decomposition of 2D Affine Transformation}
\label{sec:app_affineDecomposition}
This subsection describes how to decompose an affine transformation using the proposed SKS and ACA methods. For SKS, the kernel transformation $\mathbf{H}_{K}$ will be affine if the homography becomes an affine transformation. Thus the parameters $\mathbf{b}_{K}$ and $\mathbf{v}_{K}$ denoting the projective components in Eq.~\ref{equ:H_K} (or the parameters $1\!-\!\mathbf{a}_{K}^{'}$ and $\mathbf{b}_{K}^{'}$ in Eq.~\ref{equ:new2_H_K}) should be zero. Then an affine transformation denoted by $\mathbf{H}_{A}$ can be decomposed as
\begin{equation}
\mathbf{H}_{A} = \mathbf{H}_{S_2}^{-1}\mathbf{H}_K\mathbf{H}_{S_1},
\label{equ:affine_SKS}
\end{equation}
with a new form of $\mathbf{H}_K$
\begin{equation}
\mathbf{H}_K \equiv 
\left[\begin{array} {ccc}
1 & u_K &  \\
     & c_K &   \\
  &  & 1 \end{array}\right].
\label{equ:ACA_H_K}
\end{equation}

Then the upper left $2$$*$$2$ sub-matrix of $\mathbf{H}_{A}$ (which denotes a 2D linear transformation) can be given by
\begin{equation}
\begin{split}
{\mathbf{H}_A}^{2*2} &=  ({\mathbf{H}_{S_2}^{-1}})^{2*2}{\mathbf{H}_K}^{2*2}\mathbf{H}_{S_1}^{2*2},
 \end{split}
\end{equation}
where this expression is actually equivalent to twice using QR factorization~\cite{Golub2013}~[p.~246].

For the ACA decomposition, it is obvious that the core transformation $\mathbf{H}_{C}$ will be an identity matrix if the homography degenerates to an affine transformation with $6$ DOF. Thus the affine transformation $\mathbf{H}_{A}$ is expressed by
\begin{equation}
\begin{split}
\mathbf{H}_{A} &= \mathbf{H}_{A_2}^{-1}\mathbf{H}_{A_1} = \mathbf{H}_{A_2}^{-1}\begin{bmatrix} 1 & & \\
& 1 &  \\ 1 & 1 & 1 \end{bmatrix}\begin{bmatrix} 1 & & \\
& 1 &  \\ -1 & -1 & 1 \end{bmatrix}\mathbf{H}_{A_1} \\
&=\begin{bmatrix} N_2.x & P_2.x & M_2.x \\
N_2.y & P_2.y & M_2.y \\ 1 & 1 & 1 \end{bmatrix}\begin{bmatrix} N_1.x & P_1.x & M_1.x \\
N_1.y & P_1.y & M_1.y \\ 1 & 1 & 1 \end{bmatrix} ^{-1}.
\end{split}
\label{equ:AffineACA}
\end{equation}

\vspace{-2mm}
The second row in the above equation is actually the solution of $\mathbf{H}_{A}$ directly utilizing the constraints of the three point correspondences~\cite{ComputerGraphics}~[Eq.~10.73 in p.~237]. However, both the first and second matrices in the second row are projective transformations, which should not appear in an affine decomposition. 
Another advantage of the proposed ACA decomposition in the first row is that affine transformations can be computed in the same way as projective transformations.
Moreover, referring to Table~\ref{tab:FLOPs_ACA_4pts}, it is not difficult to deduce that the FLOPs of $\mathbf{H}_{A}=\mathbf{H}_{A_2}^{-1}\mathbf{H}_{A_1}$ is only $33$, which is obviously less than the FLOPs of the second row. Compared to the OpenCV's function \textit{getAffineTransform} which solves a linear system of inhomogeneous equations with $6$ variables utilizing the LU factorization ($\sim$\!$1000$ FLOPs), the ACA decomposition for solving an affine transformation with three point correspondences is much more efficient.
\begin{figure}[t]
\begin{center}
\includegraphics[width=0.49\textwidth]{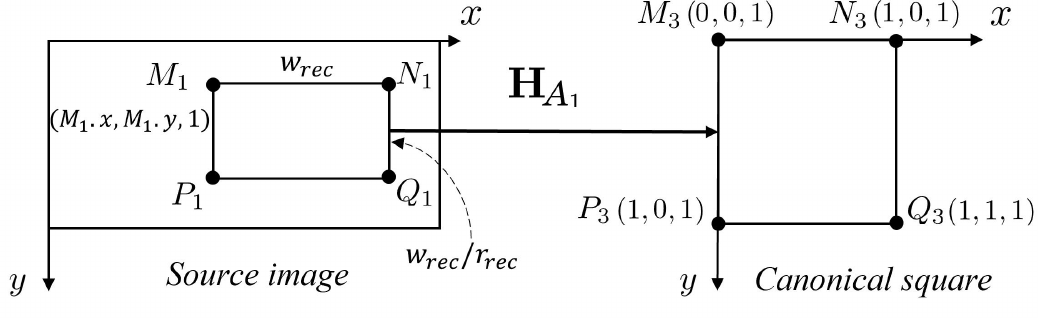}
\vspace{-8mm}
\caption{Affine transformation mapping a rectangle (with four parameters) in source image to the canonical square.}
\label{fig:Rectangle_ACA}
\end{center}
\vspace{-6mm}
\end{figure}

\vspace{-2mm}
\subsection{Tensorized ACA}
\label{sec:app_tensorACA}
\vspace{-1mm}
This subsection illustrates how to tensorize the proposed ACA decomposition for being used in deep homography pipelines. The tensorized ACA is called TensorACA, which inherits the nomination of the previous method TensorDLT~\cite{UDHN_RAL18} internally implementing the tensorized GPT-LU~\cite{OpenCV_GPT}. Although ACA only uses basic arithmetic operations which are naturally differential, two adjustments should be made to promote running efficiency.

The first adjustment is that when writing Python algorithms under deep learning frameworks, such as PyTorch and TensorFlow, it is better to take advantage of tensor structures (such as vectors, 2D matrices and 3D tensors) and operations.
By doing this, although the number of FLOPs of the adjusted algorithms increases, the runtime decreases as the cost of function calling and memory access is reduced, especially when running on GPU.

The second adjustment is for the simple quadrangle selected in source image. 
All previous $4$-point offsets based deep homography methods compute the homography mapping a square~\cite{UDHN_RAL18, DHDS_CVPR20, LocalTrans_ICCV21, DAMG_TCSVT22, IDHN_CVPR22} or rectangle~\cite{UDIS_TIP21}~\cite{CAUDHN_ECCV20} in source image to a general quadrangle in target image. However, none of these methods explore how to simplify the $4$-point homography computation under this rectangle scenario.
For a square or rectangle on source image, we find that $4$-point homography computation can be simplified as follows.

Following the image coordinate system, we assume $M_1$ is the upper left vertex of an arbitrary rectangle and $N_1$, $Q_1$, $P_1$ are other vertices in a clockwise order. As shown in Fig.~\ref{fig:Rectangle_ACA}, the affine transformation $\mathbf{H}_{A_1}$ induced by three anchor points $\{M_1,N_1,P_1\}$ transforms the rectangle to the canonical square. Denote the rectangle by four common parameters $\{M_1.x, M_1.y, w_{rec}, r_{rec}\}$, where $w_{rec}$ and $r_{rec}$ are the width and the aspect ratio respectively. The affine transformation $\mathbf{H}_{A_1}$ in Eq.~\ref{equ:AffineExpresion} is simplified to
\begin{equation}
\begin{split}
\mathbf{H}_{A_1} \equiv \begin{bmatrix}
1 &  &  \\
  & r_{rec} &  \\
   &  & 1  \end{bmatrix} 
 \begin{bmatrix}
1 &  & -M_1.x \\
  & 1 & -M_1.y \\
   &  & w_{rec}  \end{bmatrix},
\end{split}
\label{equ:H_A1_square}
\end{equation}
and the point $Q_3$ in Eq.~\ref{equ:Q3} is directly obtained without any calculation,
\begin{equation}
\begin{bmatrix} Q_3.x \\
Q_3.y \\ f_{A_1} \end{bmatrix} \equiv  \begin{bmatrix} 1 \\
1 \\ 1 \end{bmatrix}.
\label{equ:Q3_square}
\end{equation}

Substituting into Eq.~\ref{equ:t1} and Eq.~\ref{equ:C_11}, the core transformation $\mathbf{H}_{C}$ is expressed by
\begin{equation}
\mathbf{H}_C \equiv \begin{bmatrix} Q_4.x & & \\
& Q_4.y &  \\ Q_4.x\!+\!t_2 & Q_4.y\!+\!t_2 & -t_2 \end{bmatrix}.
\label{equ:H_C_square}
\end{equation}

After some manipulations, the homography $\mathbf{H}$ between a rectangle and its corresponding quadrangle is expressed by 
\begin{equation}
\begin{split}
\mathbf{H} \equiv &\begin{bmatrix} N_2.x & P_2.x & M_2.x \\
N_2.y & P_2.y & M_2.y \\ 1 & 1 & 1 \end{bmatrix} *\\
&\begin{bmatrix} Q_4.x & & -Q_4.xM_1.x  \\
& r_{rec}Q_4.y & -r_{rec}Q_4.yM_1.y  \\ t_2 & r_{rec}t_2 & -t_2(M_1.x+r_{rec}M_1.y+w_{rec}) \end{bmatrix},
\end{split}
\label{equ:H_square}
\end{equation}
where the expression will be further simplified when the rectangle is a square (i.e., $r_{rec}\!=\!1$).\begin{table*}[t]
\begin{center}
\caption{Average runtime ($\upmu$s) of different $4$-point homography algorithms running on a single CPU core under three compiler optimization options. The suffix (*) and (**) denote the implementations with single-precision and double-precision floating-point numbers, respectively. The 1st and 2nd fastest results are highlighted in red and blue, respectively.}
\label{tab:runtime_singleHomo_CPU}
\vspace{3mm}
\resizebox{0.99\textwidth}{!}{
\begin{tabular}{|l||c|c|c|c||c|c|c|c|}
\hline
\multirow{2}{*}{\tabincell{l}{Compiler \\ Optimization}} & \multicolumn{4}{c||}{Previous Methods} & \multicolumn{4}{c|}{Our Methods} \\
\cline{2-9}
 & \begin{scriptsize}NDLT-SVD(**)~\cite{Hartley2003Multiple}\end{scriptsize} & \begin{scriptsize}HO-SVD(**)~\cite{HO_BMVC05}\end{scriptsize} & \begin{scriptsize}GPT-LU(**)~\cite{OpenCV_GPT}\end{scriptsize} & \begin{scriptsize}RHO-GE(*)~\cite{Bazargani2015Fast}\end{scriptsize} & \begin{scriptsize}SKS(*)\end{scriptsize} & \begin{scriptsize}SKS(**)\end{scriptsize} &  \begin{scriptsize}ACA(*)\end{scriptsize} & \begin{scriptsize}ACA(**)\end{scriptsize} \\
\hline
Od & 13.6 & 13.5 & 0.885 & 0.169 & 0.0595 & 0.0901 & \textcolor{red}{\textbf{0.0393}} & \textcolor{blue}{\textbf{0.0561}} \\
\hline
O1 & 12.7 & 12.2 & 0.735 & 0.0327 & 0.0253 & 0.0263 & \textcolor{red}{\textbf{0.0146}} & \textcolor{blue}{\textbf{0.0178}} \\
\hline
O2 & 12.5 & 12.2 & 0.732 & 0.0287 & 0.0252 & 0.0256 & \textcolor{red}{\textbf{0.0145}} & \textcolor{blue}{\textbf{0.0171}} \\
\hline
\end{tabular}
}
\end{center}
\vspace{-5mm}
\end{table*}

It is clear that all calculation steps related to $\mathbf{H}_{A_1}$ in Table~\ref{tab:FLOPs_ours_4pts} are removed and matrix multiplication is also simplified. FLOPs of homography computation for a rectangle and square on source plane is only $47$ and $44$, respectively. As a result, the runtime of TensorACA applied in deep homography pipeline will further decrease. The complete steps of the tensorized ACA for a rectangle are illustrated in Algorithm~\ref{alg:ACA_square} with only $15$ vector operations.
It is also straightforward to see that the first two columns of homography constraining intrinsic parameters used in camera calibration~\cite{Zhang_PAMI00} are only related to the aspect ratio of the source rectangle, rather than its position and width.

It is mentioned in Sec.~\ref{sec:relatedWorks_LinearSystem} that RHO-GE without pivoting chosen may not be robust. This problem becomes more apparent when facing the four vertices of a rectangle in source image. When the four point correspondences are given in an improper order, RHO-GE will fail to compute the correct homography. A detailed discussion of failure cases is given in Appendix~\ref{appendix:C}.
\begin{algorithm}[t]
\caption{Pseudo-code of TensorACA for a rectangle}
\label{alg:ACA_square}
\begin{footnotesize}
\KwIn{rectangle's upper left vertex $M_1$, width $w_{rec}$ and aspect ratio $r_{rec}$ in source image; \\
\quad \quad \quad  four projection corners $M_2$, $N_2$, $P_2$, $Q_2$ in target image \\ \quad \quad \quad with homogeneous vector representations $\mathbf{m}_{2}$, $\mathbf{n}_{2}$, $\mathbf{p}_{2}$, $\mathbf{q}_{2}$.}

\KwOut{2D homography matrix $\mathbf{H}\!=\![\mathbf{h}_{1}, \mathbf{h}_{2}, \mathbf{h}_{3}]$ up to a scale.} 

1. Calculate two difference vectors: \\
\quad \quad $\mathbf{d}_{1} = [N_2.x\!-\!M_2.x;P_2.x\!-\!M_2.x;Q_2.x\!-\!M_2.x]$, \\ 
\quad \quad $\mathbf{d}_{2} = [N_2.y\!-\!M_2.y;P_2.y\!-\!M_2.y;Q_2.y\!-\!M_2.y]$.

2. Calculate cross-product: \\
\quad \quad $\mathbf{c} = \mathrm{Cross}(\mathbf{d}_{1},\mathbf{d}_{2})$.   \quad\quad \%  $\mathbf{c}=[Q_4.x; Q_4.y; -f_{A_2}]$

3. Calculate a temporary vector by scaling $\mathbf{m}_2$: \\
\quad \quad $\mathbf{b} = \mathrm{Sum}(\mathbf{c})\,.\!*\mathbf{m}_2$.  \quad\quad \%  $\mathrm{Sum}(\mathbf{c})=-t_2$

4. Compute the first two columns of $\mathbf{H}$: \\
\quad \quad $\mathbf{h}_1 = Q_4.x\,.\!*\mathbf{n}_2-\mathbf{b}$, \\
\quad \quad $\mathbf{h}_{2} = r_{rec}.\!*(Q_4.y\,.\!*\mathbf{p}_2-\mathbf{b}$).  \quad\quad \% $r_{rec}\!=\!1$ for a square

5. Compute the last column of $\mathbf{H}$: \\
\quad \quad $\mathbf{h}_3 = w_{rec}\,.\!*\mathbf{b}- M_1.x\,.\!*\mathbf{h}_1- M_1.y\,.\!*\mathbf{h}_2$.

\noindent \textbf{Notation:} `$.*$' represents a vector multiplied by a scalar.
\end{footnotesize}
\end{algorithm}

\vspace{-2mm}
\section{Experiments}
\label{sec:experiments}
This section shows the experimental results. As the proposed SKS and ACA deal with the task of homography computation under the minimal condition, our experiments focus on verifying the runtime of $4$-point homography solvers under various scenarios. 
Specifically, Sec.~\ref{sec:exp_CPU} tests the runtime of single homography on CPU. Sec.~\ref{sec:exp_RANSAC} further gives the changes of runtime after integrating the proposed method into the traditional feature-based RANSAC pipelines on synthetic and real datasets. Sec.~\ref{sec:exp_deeplearning} tests the runtime of different $4$-point homography solvers implemented by Python, as well as the performance in deep homography pipelines. Sec.~\ref{sec:exp_GPU} further gives the runtime of parallel computation of multiple $4$-point homographies on GPU utilizing the CUDA toolkit~\cite{NVIDIA_CUDA}. 
Detailed hardware and software configurations are listed below.

\quad \textit{CPU:} Intel i7-10700 (8 physical cores) 

\quad \textit{GPU:} NVIDIA RTX-3090 (10496 CUDA cores)

\quad \textit{Memory:} 64G 

\quad \textit{Software Environment (C++):} Win10 \& Visual Studio 2019  \& OpenCV 4.5.1 \& CUDA 11.8

\quad \textit{Software Environment (Python):} Ubuntu 20.04 \& PyTorch 1.13.0 \& CUDA 11.1

\textbf{All the shown results can be reproduced with our source codes, which have been released in \url{https://github.com/cscvlab/SKS-Homography}.}

\vspace{-2mm}
\subsection{Runtime of Single Homography on CPU}
\label{sec:exp_CPU}
In this subsection, four simulated point correspondences are used to test the runtime of homography solvers on a single core of CPU. 
Except the proposed SKS and ACA methods, there are other four algorithms participating in the comparison of $4$-point homography computation. The implementation functions and files in OpenCV 4.5.1 of these four algorithms are listed below:
\begin{itemize}
\item NDLT-SVD~\cite{Hartley2003Multiple}: \textit{runKernel} in `fundam.cpp'. 
\item HO-SVD~\cite{HO_BMVC05}: \textit{homographyHO} in `ippe.cpp'.
    \item GPT-LU~\cite{OpenCV_GPT}: \textit{getPerspectiveTransform} in `imgwarp.cpp'.
    \item GE-RHO~\cite{Bazargani2015Fast}: \textit{hFuncRefC} in `rho.cpp'.
\end{itemize}

Similar to the function \textit{hFuncRefC} in the RHO-GE method, we implement SKS and ACA without any externally defined functions or data structures. 
Meanwhile, two versions of our methods are implemented using single-precision floating point numbers (as RHO-GE uses) and double-precision floating point numbers (as NDLT-SVD, HO-SVD and GPT-LU use), respectively.  
These two implementations are denoted with the suffix (*) and (**), respectively. 

The release program runs on a single core of CPU with the compiler optimization options Od, O1 and O2, respectively~\cite{microsoftCompiler}. Optimization options O1 and O2 set a combination of optimizations including vectorization~\cite{IntelCompiler}~[p.~18-83] that generate minimum size code and optimize code for maximum speed, respectively. 
Vectorization is actually a special case of single instruction multiple data (SIMD), operating on multiple data in parallel~\cite{IntelCompiler}[p.~3-34].
On the contrary, the Od option turns off all compiler optimizations and executes program code one by one.

The average runtime is shown in Table~\ref{tab:runtime_singleHomo_CPU}, where we run all algorithms $10$M times.
Several observations are achieved.
First, the runtime difference of using single-precision and double-precision floating point numbers is not significant, especially when enabling compiler optimization (O1 or O2).
Second, the influence of compiler optimization for the algorithms differs obviously. For the first three algorithms from left to right based on OpenCV data structures and functions, the influence of enabling compiler optimization is trivial. However, for the last five algorithms (RHO-GE and ours) including only arithmetic operations, doing this brings several times speedup.
Of these algorithms, RHO-GE benefits the most from enabling compiler optimizations as it is pointed out that `This neatly fits in half of a vector register file with 16 4-lane registers, a common configuration in most modern architectures.'~\cite{Bazargani2015Fast}.
Third, compared to the three robust methods (NDLT-SVD, HO-SVD and GPT-LU), SKS(**) and ACA(**) represents \{$488$\textit{x}, $477$\textit{x}, $29$\textit{x}\} and \{$731$\textit{x}, $713$\textit{x}, $43$\textit{x}\} respectively under O2 optimization. 
These values of practical speedup are significantly larger than the theoretical FLOPs comparison shown in Table~\ref{tab:relatedWork_pointsHomography}.
This is because the implementations of the three robust methods include more or less conditional branch judgments, data copy or exchange, OpenCV data structures, etc., which severely influence the speed.
This phenomenon will be alleviated when we run these algorithms on GPU shown in Sec.~\ref{sec:exp_GPU} as all external data structures and functions are discarded except the CUDA toolkit. 
Fourth, SKS(*) is obviously faster than RHO-GE, while ACA(*) takes only half runtime of RHO-GE under O2 optimization, representing $\sim$$70$M calculations per second. The runtime performance of SKS and ACA is basically consistent with their FLOPs. 
This speedup is a significant improvement for estimating homography between images, whether in the traditional feature-based RANSAC pipelines tested in Sec.~\ref{sec:exp_RANSAC}, or in deep homography pipelines tested in Sec.~\ref{sec:exp_deeplearning}. 

\begin{table*}[tb!]
\begin{center}
\caption{Average runtime ($\upmu$s) of five stages for six combinations of feature points extraction and matching frameworks on MS-COCO dataset. 
The 1st and 2nd fastest results of computing $4$-point homography are highlighted in red and blue, respectively. The notations `$\mathcal{R}$' and `$\mathcal{M}$' denote the default solver in RANSAC and MAGSAC++ respectively.}
\label{tab:runtime_RANSAC_MSCOCO}
\vspace{3mm}
\resizebox{0.99\textwidth}{!}{
\begin{tabular}{|l|l||c|c|c|c|c|c|}
\hline
\multicolumn{2}{|l||}{Stage} & \tabincell{c}{\begin{scriptsize}SIFT~\cite{SIFT2004} \&\end{scriptsize} \\ \begin{scriptsize}RANSAC~\cite{fischler_ACMC81}\end{scriptsize}} & \begin{scriptsize}\tabincell{c}{SIFT~\cite{SIFT2004} \& \\ MAGSAC++~\cite{MAGSACplusplus}}\end{scriptsize} & \begin{scriptsize}\tabincell{c}{ORB~\cite{ORB} \& \\ RANSAC~\cite{fischler_ACMC81}}\end{scriptsize} & \begin{scriptsize}\tabincell{c}{ORB~\cite{ORB} \& \\ MAGSAC++~\cite{MAGSACplusplus}}\end{scriptsize} & \begin{scriptsize}\tabincell{c}{SuperPoint~\cite{SuperPoint18} \& \\ RANSAC~\cite{fischler_ACMC81}}\end{scriptsize} & \begin{scriptsize}\tabincell{c}{SuperPoints~\cite{SuperPoint18} \& \\ MAGSAC++~\cite{MAGSACplusplus}}\end{scriptsize} \\
\hline
\multicolumn{2}{|l||}{1)\; Feature Points Extraction} & \multicolumn{2}{c|}{19.7K} & \multicolumn{2}{c|}{4.71K} & \multicolumn{2}{c|}{12.2K } \\
\hline
\multicolumn{2}{|l||}{2)\; Descriptor Matching w. Ratio Test} & \multicolumn{2}{c|}{680} & \multicolumn{2}{c|}{1.20K} & \multicolumn{2}{c|}{17.2K} \\
\hline
\multirow{6}{*}{\tabincell{l}{3)\; Loop of \\ $4$-point \\ Homography \\ Computation}} & NDLT-SVD~\cite{Hartley2003Multiple} ($\mathcal{R}$)  & 109 & 76.8 & 204 & 154 & 397 & 314 \\
\cline{2-8}
 & HO-SVD~\cite{HO_BMVC05}  & 79.7 & 56.3 & 144 & 98.7 & 279 & 215 \\
 \cline{2-8}
 & GPT-LU~\cite{OpenCV_GPT} ($\mathcal{M}$)  & 8.03  & 4.72 & 14.7 & 8.46 & 29.8 & 16.3 \\
 \cline{2-8}
 & RHO-GE~\cite{Bazargani2015Fast}  & 0.341 & 0.217 & 0.503 & 0.400 & 1.24  & 0.771  \\
 \cline{2-8}
 & SKS (\textbf{Ours})  & \textcolor{blue}{\textbf{0.238}} & \textcolor{blue}{\textbf{0.191}} & \textcolor{blue}{\textbf{0.422}} & \textcolor{blue}{\textbf{0.331}} & \textcolor{blue}{\textbf{0.887}} & \textcolor{blue}{\textbf{0.676}} \\
 \cline{2-8}
 & ACA (\textbf{Ours}) & \textcolor{red}{\textbf{0.174}} & \textcolor{red}{\textbf{0.144}} & \textcolor{red}{\textbf{0.302}} & \textcolor{red}{\textbf{0.281}} & \textcolor{red}{\textbf{0.690}} & \textcolor{red}{\textbf{0.517}} \\
\hline
\multicolumn{2}{|l||}{4)\; Loop of Verification} & 4.34 & 25.6 & 6.41 & 42.0 & 19.9 & 153 \\
\hline
\multicolumn{2}{|l||}{5)\; Global Optimization} & 101 & 54.3 & 82.0 & 79.4 & 108 & 103 \\
\hline
\multirow{4}{*}{\tabincell{l}{Metrics}} & 
Inlier Number & 204 & 202 & 137 & 135 & 224 & 225 \\ 
\cline{2-8}
 & Inlier Ratio & 0.919 & 0.912 & 0.846 & 0.833 & 0.731 & 0.735 \\
 \cline{2-8}
 & Homography Loop  & 5.68 & 3.51 & 9.97 & 7.27 & 18.8 & 15.4 \\
 \cline{2-8}
 & Total Runtime w. $\mathcal{R}$/$\mathcal{M}$ & 20.6K & 20.5K & 6.20K & 6.04K & 29.9K & 29.7K \\
\hline
\end{tabular}
}
\end{center}
\vspace{-5mm}
\end{table*}

\vspace{-2mm}
\subsection{Runtime in Feature-based RANSAC Pipelines}
\label{sec:exp_RANSAC}
\vspace{-1mm}
In this subsection, SKS and ACA are integrated as a plug-in module into the traditional feature-based RANSAC pipelines to estimate homography between two images.
The main purpose of this test is to observe how our homography decomposition methods accelerate practical applications. 
For the comprehensive comparison of feature-based RANSAC pipelines, interested readers could refer to the literature focusing on feature points extraction and matching, such as the  benchmark~\cite{ImageMatching_IJCV21}.

The first experiment is conducted on the widely used synthetic MS-COCO dataset~\cite{MSCOCO}.  Following the early work~\cite{arXiv16}, 5000 images are randomly chosen from MS-COCO's test set.
Each image is resized to grayscale $640$*$480$, and a pair of $256$*$256$ image patches is generated by a synthetic homography, which matches four corners of a source square patch to a target quadrangle patch with four disturbed corners. The whole process is roughly divided into five stages, which are tested separately.
The first two stages accomplish the feature points extraction and coarse matching. 
In the first stage, the traditional SIFT~\cite{SIFT2004} and ORB~\cite{ORB} algorithms, as well as the deep learning method SuperPoint~\cite{SuperPoint18}, are used to extract feature points whose maximum number is set to $1000$. 
In the second stage, coarse descriptor matching based on $2$ nearest neighbors search and ratio test (with the threshold $\tau$$=$$0.8$) is performed. Image pairs with matching feature points less than 8 are removed.
Then, RANSAC~\cite{fischler_ACMC81} and the SOTA MAGSAC++~\cite{MAGSACplusplus} frameworks, covering the third to fifth stages, are adopted to remove outliers. 
The third stage is the loop computation of $4$-point homography, where the previous four solvers and our two solvers participate in the comparison.
Notice that the default $4$-point homography solver in RANSAC and MAGSAC++ is NDLT-SVD and GPT-LU respectively.
The fourth stage includes loop of $4$-point subset sample, inliers verification, and other possible operations (e.g., shape check of $4$-point, homography scoring and local optimization in MAGSAC++).
The number of loops here is calculated based on inlier ratio with a probability $99\%$ that all sampled four points are inliers. At the same time, the maximum number of loops is set to 1000, roughly corresponding to the minimum inlier ratio of $26\%$.
The fifth stage is the global optimization with all inliers, in which the Levenberg-Marquardt (LM) algorithm~\cite{LM} and a series of manipulations are adopted in RANSAC and MAGSAC++ respectively.
All the above methods are implemented based on OpenCV's C++ procedures and tested on CPU, except SuperPoint which is implemented using PyTorch and tested on GPU.
The average runtime of six combinations of three feature points methods and two outlier-removal frameworks is shown in Table~\ref{tab:runtime_RANSAC_MSCOCO}.

Overall speaking, similar to the results shown in Table~\ref{tab:runtime_singleHomo_CPU}, 
SKS and ACA significantly reduce runtime of the stage of $4$-point homography computation.
However, compared to the default robust solvers NDLT-SVD and GPT-LU, the speedup brought by SKS or ACA for the whole process is not obvious as the first two stages of feature points extraction and coarse matching are relatively time-consuming.
Another reason is that the synthetic image pairs have high inlier ratios (shown in the second row of `Metrics') and small numbers of homography loops (shown in the third row of `Metrics').
For example, the maximum speedup happens in the combination of the real-time feature point method ORB and the classical RANSAC framework. Replacing the default solver NDLT-SVD with our ACA will save about $204$ microseconds, which is accumulated from $7.27$ loops of $4$-point homography computation.
Compared to the original process with the total runtime $6.20$K, the revised process represents about $3\%$ speedup and $97\%$ runtime ratio.
Another minor factor affecting the runtime is the number of involved points in each stage. Take the runtime of the fifth stage as an instance. Both in the RANSAC and MAGSAC++ frameworks, SuperPoint is the most time-consuming feature point method as it has the largest number of inlier points (shown in the first row of `Metrics').
\begin{figure}[t]
\begin{center}
\includegraphics[width=0.48\textwidth]{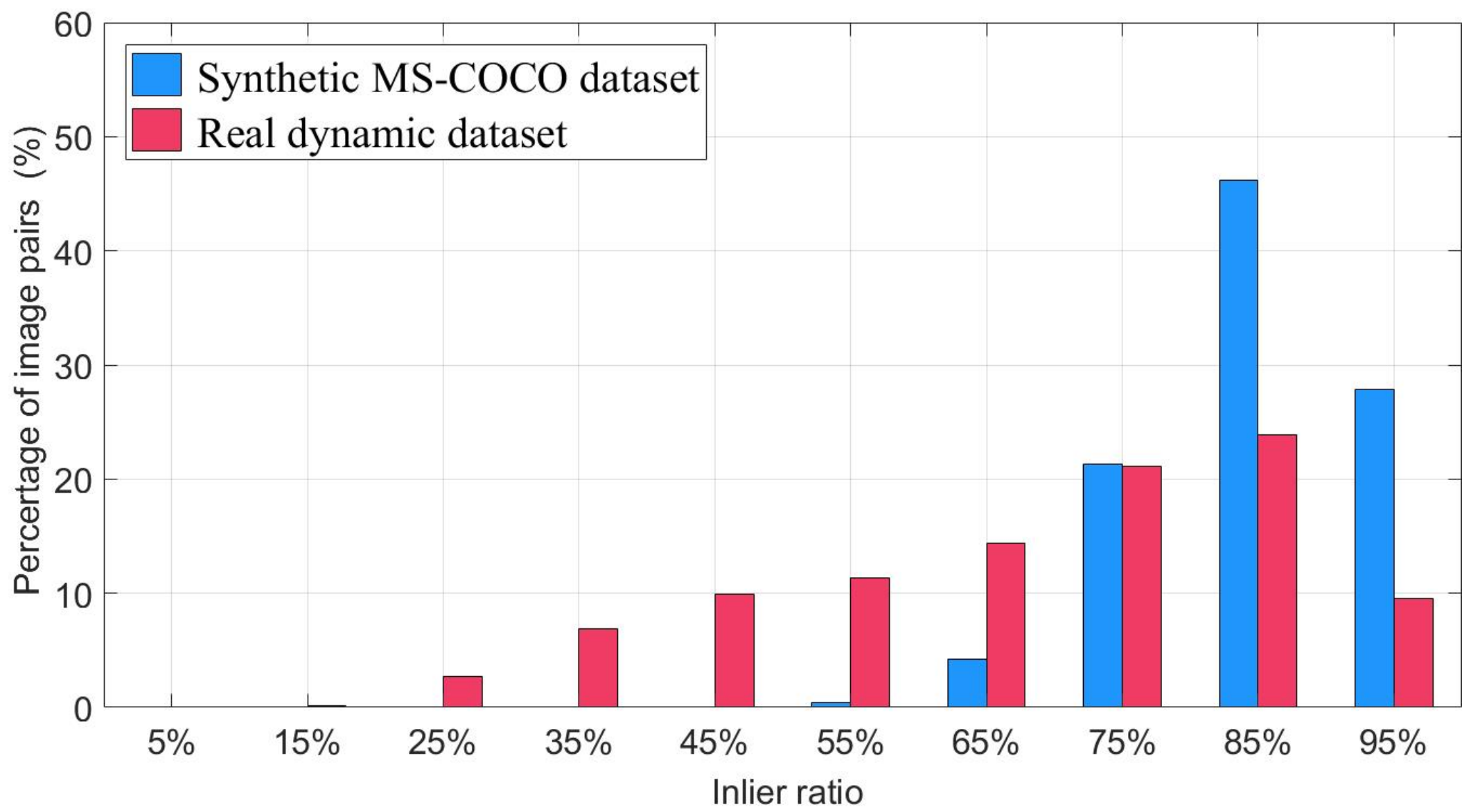}
\includegraphics[width=0.48\textwidth]{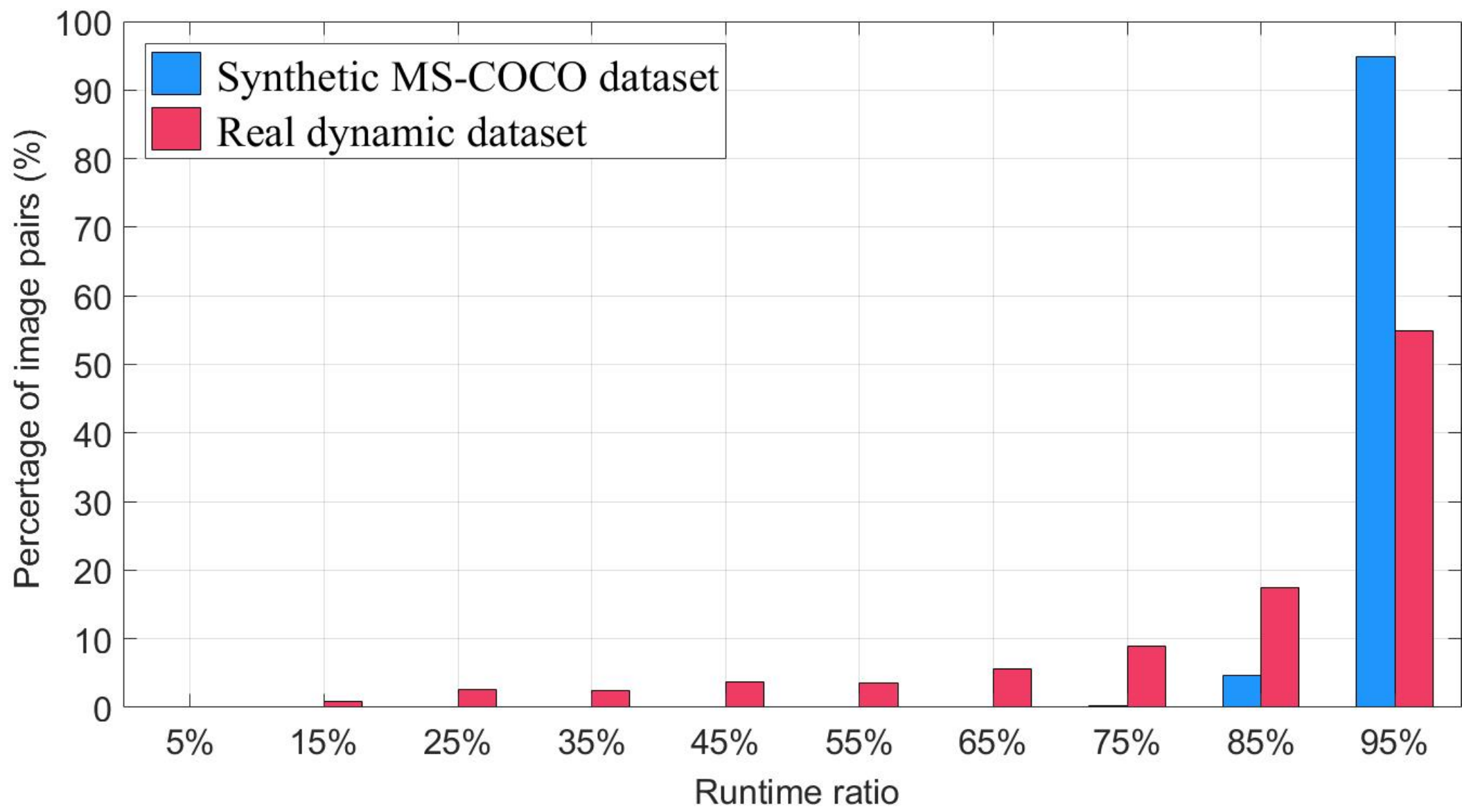}
\vspace{-3mm}
\caption{Comparison between the synthetic MS-COCO dataset and the dynamic real dataset. Upper: inlier ratio; Lower: runtime ratio.}
\label{fig:hist_inlierRatio}
\end{center}
\vspace{-8mm}
\end{figure}

Most image pairs in the Warped MS-COCO dataset are ideal in terms of lighting, texture, and no foreground, which result in their generally high inlier ratios.
Thus the second experiment is conducted on one more difficult real dataset: dynamic scenes~\cite{DHDS_CVPR20}.
From the real dynamic dataset, we randomly select $33$ videos, containing $2024$ image pairs.
The combination `ORB \& RANSAC' adopted by ORB-SLAM3~\cite{ORB-SLAM3} is performed to compare this real dataset and MS-COCO.
The runtime performance is evaluated through the processes integrating ACA and the default $4$-point homography solver NDLT-SVD.
The distribution of inlier ratio and runtime ratio (percentage of image pairs for ten levels of ratio) are shown in Fig.~\ref{fig:hist_inlierRatio}.
It is clear that the dynamic real dataset has significantly lower inlier ratio and runtime ratio than MS-COCO.
For the other two combinations containing the RANSAC framework, their runtime ratios will increase due to the increasing runtime of feature extraction and numbers of loops.
For the combinations containing the MAGSAC++ framework, the speedup brought by ACA will be further insignificant, since the default $4$-point homography solver GPT-LU already saves most of the runtime of the third stage. 

\vspace{-2mm}
\subsection{Runtime in Deep Homography Pipelines}
\label{sec:exp_deeplearning}
\vspace{-1mm}
In this section, we test the runtime of deep homography networks using either the default $4$-point homography solver TensorDLT or the proposed TensorACA.
Different from the traditional feature-based RANSAC pipelines, deep homography pipelines do not suffer the inconsistent runtime mainly caused by the undetermined number of loops. 
TensorDLT (which actually is the tensorized GPT-LU) is implemented differently in recent deep homography networks, including UDHN~\cite{UDHN_RAL18}, CA-UDHN~\cite{CAUDHN_ECCV20} and $1$-scale IDHN~\cite{IDHN_CVPR22}. 
In the official implementation using PyTorch, UDHN and CA-UDHN use the same code (represented by TensorDLT-1) to compute a $4$-point homography, although the resolutions of their input gray images are $128$$*$$128$ and $560$$*$$315$ respectively. IDHN uses another implementation code (represented by TensorDLT-2) and its input images are color with the resolution $128$$*$$128$.
TensorACA is implemented in two ways, which are called TensorACA-vanilla and TensorACA-rect. 
TensorACA-vanilla is written using Python statements to imitate our C++ program, which only contains arithmetic operations of floating-point numbers. TensorACA-rect is implemented (as shown in Algo.~\ref{alg:ACA_square}) for the mapping from a source rectangle to a target quadrangle. 
Meanwhile, we also use Python statements to re-implement the RHO-GE method named TensorGE to participate in the comparison.
For TensorGE, we manually select an order of four points analyzed in Appendix.~\ref{appendix:C} to avoid division by zero.
The two methods NDLT-SVD and HO-SVD are not re-implemented using Python as it is pointed out that `taking the gradients in SVD has high time complexity and has practical implementation issues.'~\cite{UDHN_RAL18}.

The experiments are conducted to test the inference time of the individual modules of five $4$-point homography solvers, the original networks and the revised networks both on CPU and GPU, with the PyTorch library. 
The average runtime of $50$K trials for modules and $5$K trials for networks are shown in Table~\ref{tab:runtime_compare_Tensor}. 
Several observations are illustrated below.

First, the individual module of each $4$-point homography solver runs about twice as slowly on GPU than on CPU. However, the whole network runs faster on GPU than on CPU, since GPU is suitable for large-scale parallel computing, such as convolution operations.

Second, there exists obvious difference of runtime for two ways of tensorizing GPT-LU (TensorDLT-1 and TensorDLT-2). For Python programs, it is not strange because that different operation ways for high-dimensional arrays will bring about large speed differences. Moreover, the performance of $4$-point homography solver in individual module test is a little different from integrating it in networks. This is because GPU executions are commonly asynchronous, but the stable runtime test requires synchronization~\cite{NVIDIA_CUDA_reference}.

Third, TensorGE is the slowest in the tests of both individual module and the three integrated networks, because it is inefficient for Python statements to execute floating-point arithmetic operations.
TensorACA-vanilla performs much better than TensorGE due to less FLOPs and parameters. Compared with TensorDLT-2, both of the two TensorACA methods run faster in individual module and the three integrated networks. Compared with TensorDLT-1, TensorACA-rect runs about $18\%$ and $15\%$ faster in individual module test on CPU and GPU respectively.
\begin{table*}[t]
\begin{center}
\caption{Average runtime (ms) of homography computation from $4$-point offsets in deep homography inference. In UDHN and CA-UDHN, the $4$-point homography solver TensorDLT-1 is executed only once, while in IDHN, its solver TensorDLT-2 is iteratively executed six times.}
\label{tab:runtime_compare_Tensor}
\vspace{3mm}
\resizebox{0.99\textwidth}{!}{
\begin{tabular}{|l||c|c|c|c||c|c|c|c|}
\hline
\multirow{2}{*}{Methods} & \multicolumn{4}{c||}{Running on CPU} & \multicolumn{4}{c|}{Running on GPU}
\\
\cline{2-9}
& Module  & UDHN~\cite{UDHN_RAL18} & CA-UDHN~\cite{CAUDHN_ECCV20} &  IDHN~\cite{IDHN_CVPR22} & Module  & UDHN~\cite{UDHN_RAL18} & CA-UDHN~\cite{CAUDHN_ECCV20} & IDHN~\cite{IDHN_CVPR22} \\
\hline
TensorDLT-1 (GPT-LU~\cite{OpenCV_GPT}) &  \textcolor{blue}{\textbf{0.1262}} &  \textcolor{blue}{\textbf{18.52}} &  \textcolor{blue}{\textbf{1229.26}} &  &  \textcolor{blue}{\textbf{0.2509}} &  \textcolor{blue}{\textbf{4.726}}  &  \textcolor{blue}{\textbf{13.52}} &   \\
\hline
TensorDLT-2 (GPT-LU~\cite{OpenCV_GPT}) & 0.4458 &  &  & 46.86 & 0.8286 &  &  & 23.84  \\
\hline
TensorGE (RHO-GE~\cite{Bazargani2015Fast}) & 1.372 &  20.04 & 1231.93  & 53.35 & 2.522 &  7.227  & 15.72 & 34.00 \\
\hline
TensorACA-vanilla (\textbf{Ours}) &  0.2976 & 18.77 & 1229.44 &  \textcolor{blue}{\textbf{45.97}} & 0.6972  & 5.186 &  13.82  & \textcolor{blue}{\textbf{23.16}}  \\
\hline
TensorACA-rect (\textbf{Ours}) & \textcolor{red}{\textbf{0.1034}} &  \textcolor{red}{\textbf{18.50}} &  \textcolor{red}{\textbf{1229.23}} & \textcolor{red}{\textbf{44.72}} & \textcolor{red}{\textbf{0.2145}} &  \textcolor{red}{\textbf{4.666}} & \textcolor{red}{\textbf{13.47}} & \textcolor{red}{\textbf{19.91}} \\
\hline
\end{tabular}
}
\end{center}
\vspace{-5mm}
\end{table*}

Fourth, with TensorACA-rect as the new $4$-point homography solver, the revised UDHN, CA-UDHN and IDHN can reduce $1.3\%$, $0.34\%$, and $17\%$ runtime respectively on GPU (shown in the right part of the last row of Table~\ref{tab:runtime_compare_Tensor}), compared to their original networks with TensorDLT inside (shown in the right part of the third and fourth rows of Table~\ref{tab:runtime_compare_Tensor}).
Both the absolute time reduction and largest percentage speedup occurs the revised IDHN,
as the adopted TensorDLT-2 is not efficient and its network iteratively predicts $4$-point offsets six times to further refine the homography.

Compared with the CPU runtime of these $4$-point homography solvers implemented by C++ (shown in Table~\ref{tab:runtime_singleHomo_CPU} and Table~\ref{tab:runtime_RANSAC_MSCOCO}), the performance of these solvers' Python procedures drops dramatically. For example, TensorDLT-1, TensorGE and TensorACA-vanilla converting C++ statements to Python statements are about $172$($\approx$$126.2/0.732$), $47.8$K($\approx$$1372/0.0287$) and $17.4$K($\approx$$297.6/0.0171$) times slower, respectively.
TensorDLT-1 gains the fewest runtime drop, as its Python implementation is suitable for tensorization (only involving stack of coefficient matrix and solving an inhomogeneous linear system with the form $\mathbf{A}_{8*8}\mathbf{x}_{8*1}$$=\mathbf{b}_{8*1}$).
Thus, to avoid runtime drop of Python programs and pursue fast running in practice, one possible solution is to implement networks directly with the C++ CUDA toolkit\footnote{One famous example we familiar is the 3D implicit representation neural network Instant-NGP~\cite{instantNGP_TOG22}, which can be trained within a number of seconds and accomplish new view synthesis inference in real time.}.

\vspace{-2mm}
\subsection{Runtime of Parallel Homographies on GPU}
\label{sec:exp_GPU}
\vspace{-1mm}
In this subsection, we test the runtime of multiple homographies computation in parallel on GPU.
This is meaningful for both the feature-based RANSAC pipelines and the deep homography pipelines.
In the traditional feature-based RANSAC pipelines, multiple $4$-point subsets can be sampled simultaneously from extracted point correspondences, followed by parallel computation of multiple homographies in the next stage.
In the deep homography pipelines implemented by Python in Sec.~\ref{sec:exp_deeplearning}, one and a number of (batch size) $4$-point homographies are calculated in network inference and training, respectively.
Therefore, compared to sequential execution on CPU and parallel execution with Python on GPU, direct implementation of multiple homographies computation in parallel using CUDA on GPU will greatly reduce the runtime.
Specifically, each $4$-point homography is assigned to one thread of GPU for computation and the program statements will be sequentially executed by a GPU CUDA core. We sample $\{1, 10, 100, 1000, 10K, 100K, 1M\}$ $4$-point subsets respectively, followed by sending them to GPU for parallel homography computation. These sampling numbers correspond to the approximate outlier proportions $\{0\%, 22\%, 54\%, 74\%, 85\%, 92\%, 95\%\}$ respectively, with a probability $99\%$ to ensure at least one $4$-point sample only consists of inliers.
For running on GPU, the previous four algorithms and our two methods are re-implemented using CUDA. There are only double-precision floating-point numbers adopted and all OpenCV data structures are discarded in all implementations.

\begin{table}[t]
\begin{center}
\caption{Average runtime ($\upmu$s) of computing multiple $4$-point homographies in parallel on GPU. All methods are re-implemented using CUDA. Here the listed sampling numbers correspond to different outlier ratios with a probability $99\%$ that all four points are inliers in standard RANSAC.}
\label{tab:runtime_GPU}
\vspace{0mm}
\resizebox{0.49\textwidth}{!}{
\begin{tabular}{|l||c|c|c|c||c|c|}
\hline
\multirow{3}{*}{\tabincell{l}{Samp. Num.\\ / Outlier Ratio}} & \multicolumn{4}{c||}{Previous Methods} & \multicolumn{2}{c|}{Our Methods} \\
\cline{2-7}
& \tabincell{c}{NDLT-SVD \\ \cite{Hartley2003Multiple}} & \tabincell{c}{HO-SVD \\ \cite{HO_BMVC05}} & \tabincell{c}{GPT-LU \\ \cite{OpenCV_GPT}} & \tabincell{c}{RHO-GE \\ \cite{Bazargani2015Fast}} & SKS & ACA \\ 
\hline
1\;/\;0\% & 469 & 55.1 & 29.6 & 4.69 & \textcolor{blue}{\textbf{4.20}} & \textcolor{red}{\textbf{3.11}} \\
\hline
10\;/\;$\sim\!22\%$ & 617 & 65.5 & 30.7 & 4.69 & \textcolor{blue}{\textbf{4.26}} & \textcolor{red}{\textbf{3.16}} \\
\hline
100\;/\;$\sim\!54\%$ & 794 & 79.3 & 31.0 & 4.74 & \textcolor{blue}{\textbf{4.31}} & \textcolor{red}{\textbf{3.19}} \\
\hline
1K\;/\;$\sim\!74\%$ & 807 & 80.8 & 30.8 & 6.17 & \textcolor{blue}{\textbf{4.83}} & \textcolor{red}{\textbf{3.20}} \\
\hline
10K\;/\;$\sim\!85\%$ & 1.35K & 135 & 50.7 & 10.1 & \textcolor{blue}{\textbf{7.45}} & \textcolor{red}{\textbf{5.26}} \\
\hline
100K\;/\;$\sim\!92\%$ & 15.0K & 1.19K & 845 & 66.7 & \textcolor{blue}{\textbf{49.9}} & \textcolor{red}{\textbf{29.3}} \\
\hline
1M\;/\;$\sim\!95\%$ & 151K & 11.2K & 8.39K & 589 & \textcolor{blue}{\textbf{436}} & \textcolor{red}{\textbf{245}} \\
\hline
\end{tabular}
}
\end{center}
\vspace{-6mm}
\end{table}

\begin{figure}[t]
\begin{center}
\includegraphics[width=0.46\textwidth]{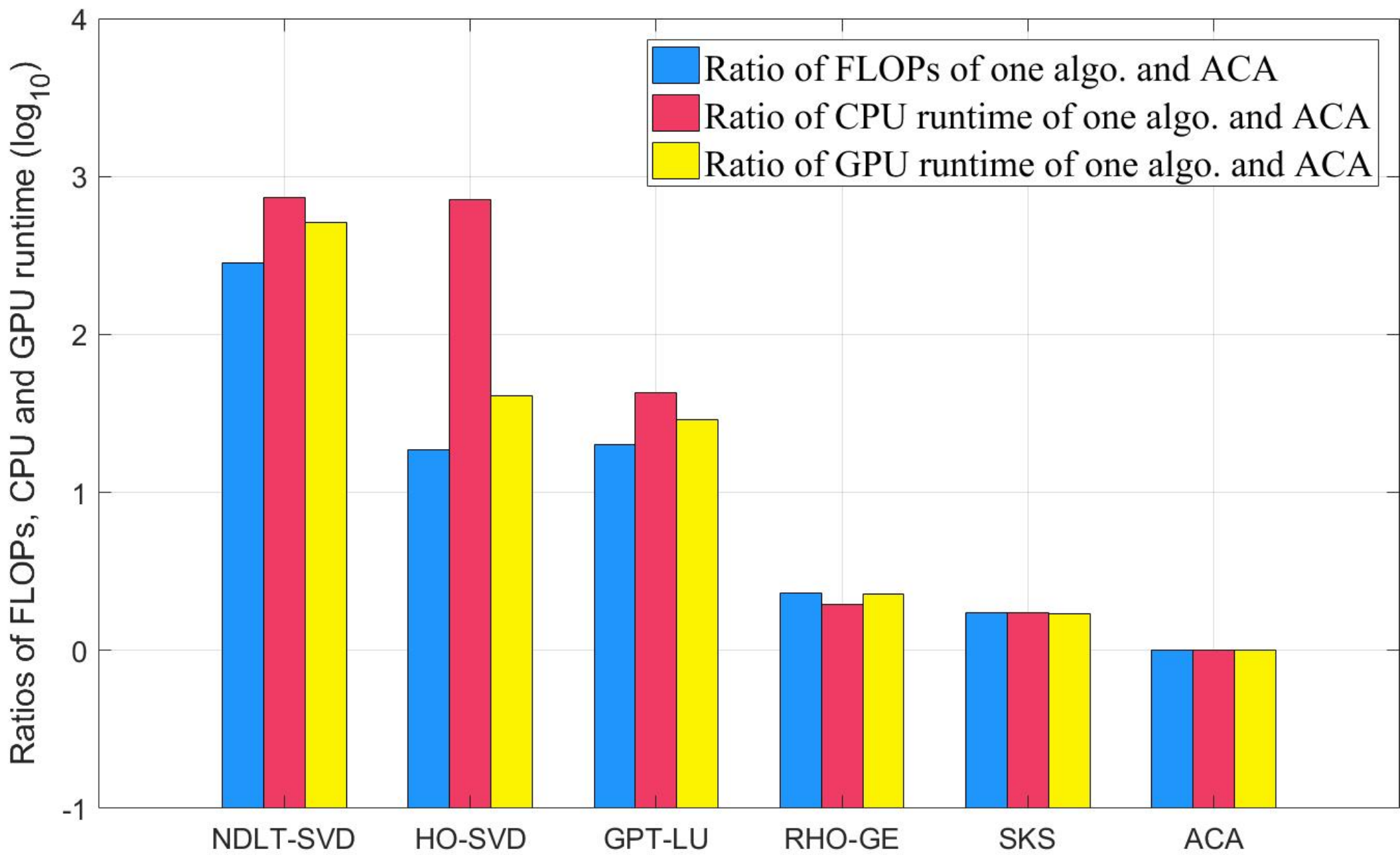}
\vspace{-3mm}
\caption{Ratios ($log_{10}$) of FLOPs, CPU runtime and GPU runtime of one algorithm and the proposed ACA as a comparison basis. Since the programs of NDLT-SVD, HO-SVD and GPT-LU running on CPU include  conditional branch judgments, data copy or exchange, OpenCV data structures, their ratios of CPU runtime deviate the ratios of FLOPs much more than the ratios of GPU runtime.}
\label{fig:FLOPs_CPU_GPU}
\end{center}
\vspace{-8mm}
\end{figure}

The average runtime of computing multiple $4$-point homographies in parallel on GPU are shown in Table~\ref{tab:runtime_GPU}, where we run the six algorithms $10$K to $1$M times, according to roughly the same total execution time ($10$ seconds). 
Overall speaking, for all sample numbers (or outlier ratios), the proposed ACA algorithm is fastest among the compared methods, followed by the proposed SKS algorithm.
The comparison of RHO-GE, SKS and ACA are basically similar to running on CPU (shown in Table~\ref{tab:runtime_singleHomo_CPU}). SKS is always obviously faster than RHO-GE, and ACA is about twice as fast as RHO-GE.
For NDLT-SVD, HO-SVD and GPT-LU re-implemented without external data structures running on GPU, their performances are more consistent with comparison of FLOPs than them on CPU.  
Taking the fastest ACA method as the basis for comparison, the ratios of FLOPs, CPU runtime (with `O2' compiler optimization) and GPU runtime (with $100$K homographies in parallel) of all algorithms are depicted in Fig.~\ref{fig:FLOPs_CPU_GPU}.
It is observed that the ratios of GPU runtime are more consistent with ratios of FLOPs for all algorithms than the ratios of CPU runtime.

\begin{figure}[t]
\begin{center}
\includegraphics[width=0.23\textwidth]{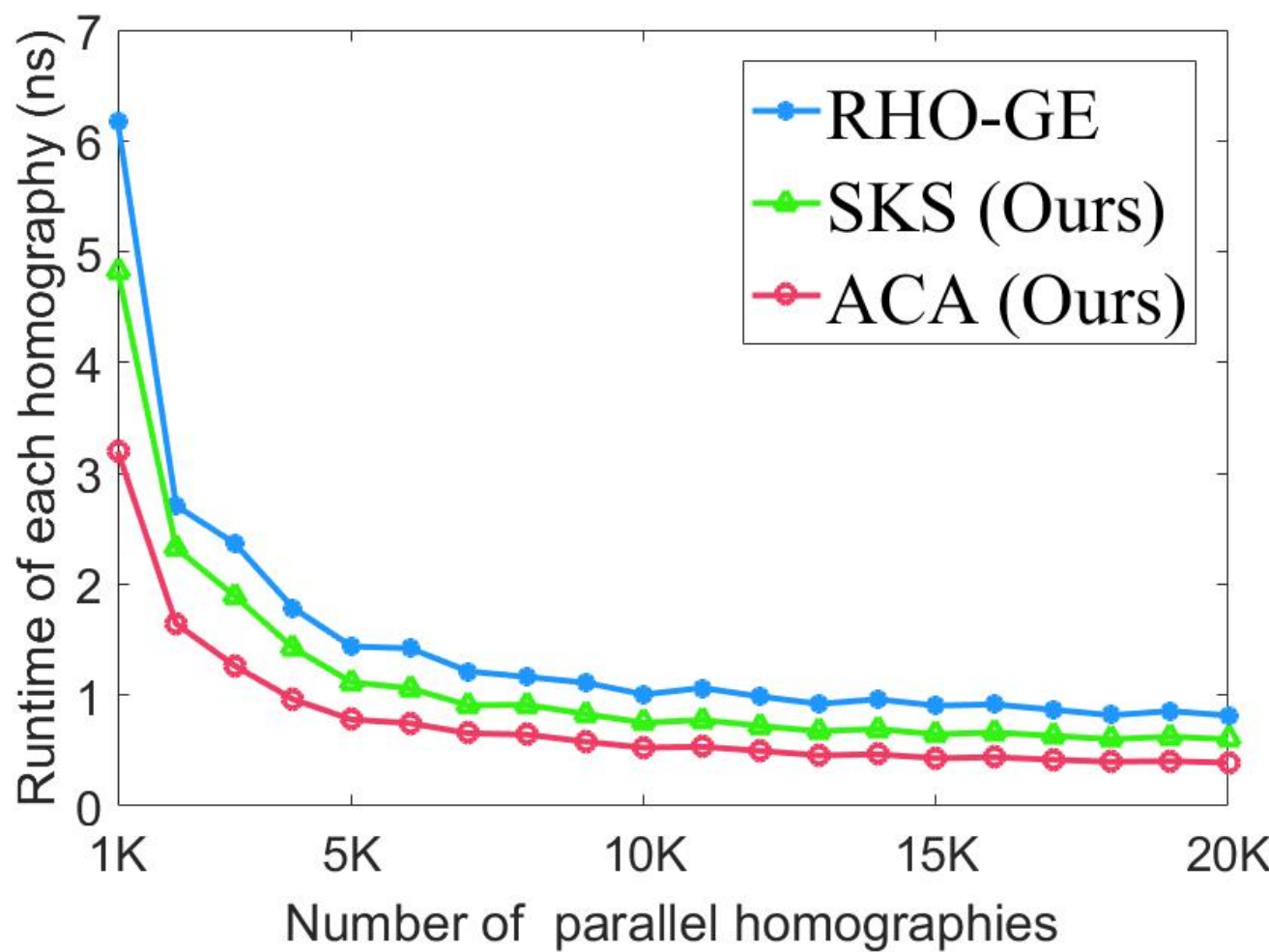}
\includegraphics[width=0.23\textwidth]{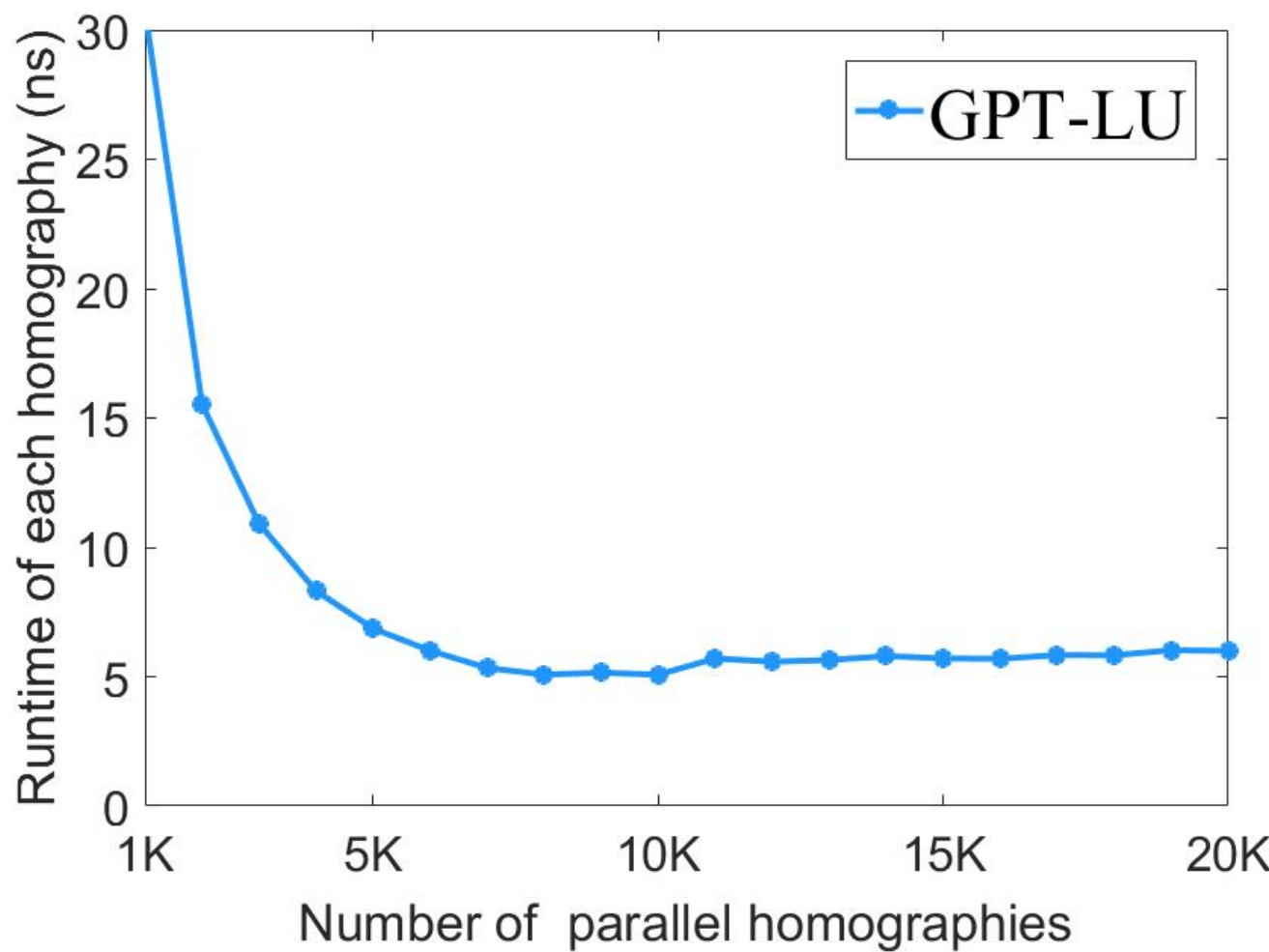}
\includegraphics[width=0.23\textwidth]{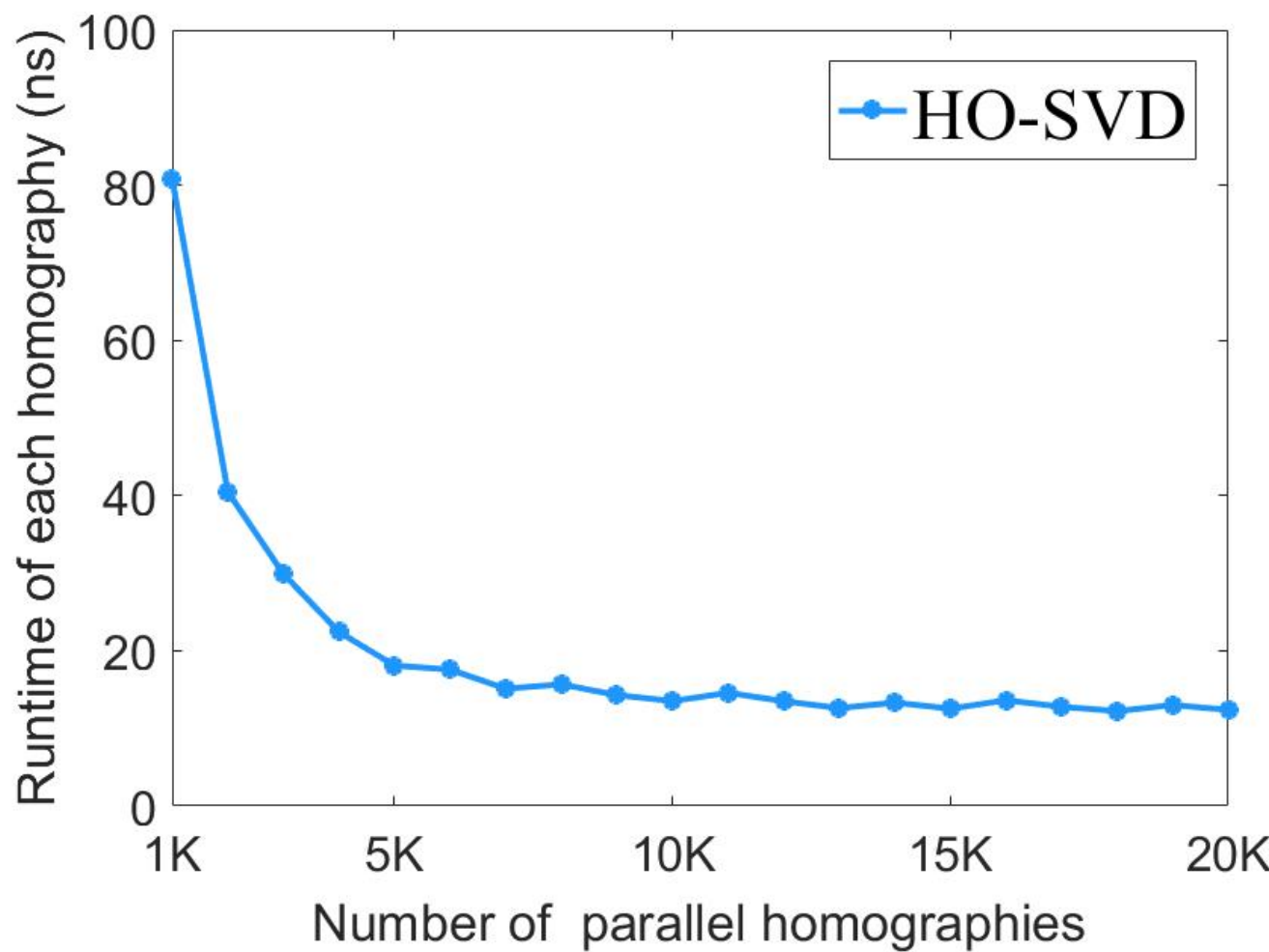}
\includegraphics[width=0.23\textwidth]{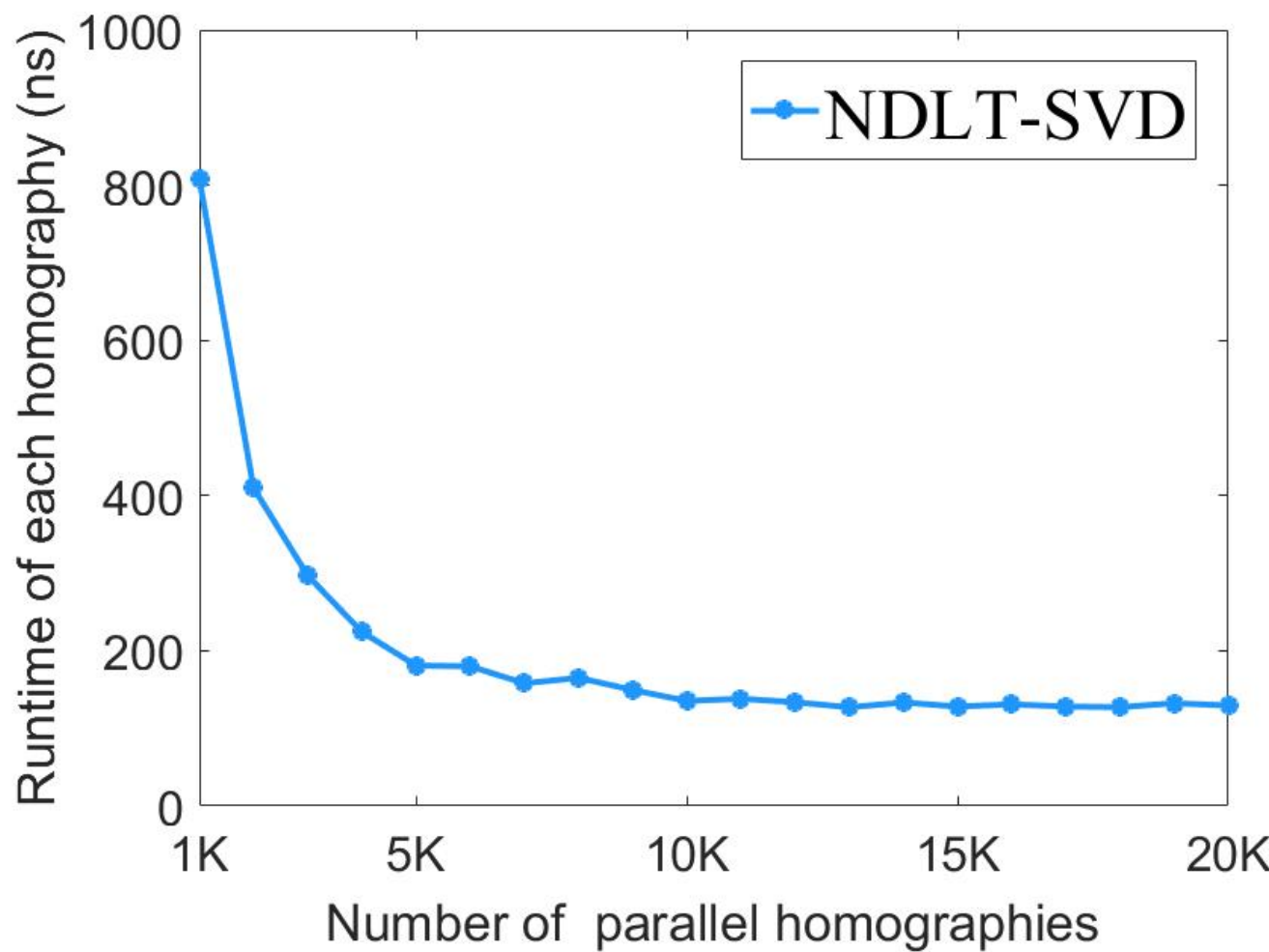}
\vspace{-3mm}
\caption{Average runtime (ns) of computing one homography on GPU. The stationary point of runtime happens with the sampling number $10$K, which is approximately equal to the number of CUDA cores (10496) of NVIDIA RTX3090 GPU.}
\label{fig:zoom_GPU}
\end{center}
\vspace{-5mm}
\end{figure}

Another observation is also expected that the total runtime of all algorithms for small numbers ($\leq$$10$K) of homographies in Table~\ref{tab:runtime_GPU} increases slightly with the increase of the numbers. Meanwhile, the average runtime of computing one homography is almost inversely proportional to the small sampling number.
This is because the parallel computation of small numbers of homographies don't trigger all $10496$ CUDA cores and the slight increase of the runtime is mainly due to more data accesses. Thus we further conduct an experiment with more sampling numbers ranging from $1$$\mathbf{K}$ to $20$$\mathbf{K}$. The results of these six methods are shown in Fig.~\ref{fig:zoom_GPU}. It can be seen that the stationary point of GPU runtime happens with the sampling number $10$K, which is the approximately same to the number of CUDA cores (10496). 
When the sampling number is higher than $10$K, the runtime of computing one homography becomes stable and the total runtime of computing multiple homographies begins to increase linearly with the increase of the number.

The efficiency difference for $4$-point homography computation on CPU and GPU can be seen from Table~\ref{tab:runtime_singleHomo_CPU} and Table~\ref{tab:runtime_GPU}. 
When the number of homographies needed to compute is less than $100$ (outlier ratio is lower than $\sim$$54\%$), the GPU runtime for one homography is actually larger than the CPU runtime. 
When the number of homographies in parallel computation increases above $100$, the efficiency advantage of GPU starts to show.
For example, the average runtime of one homography in the parallel computation of $1$M homographies on GPU is only $0.245$ns (shown in the last row of Table~\ref{tab:runtime_GPU}), representing $\sim$$70$x speedup relative to the CPU runtime $17.1$ns (shown in the last row of Table~\ref{tab:runtime_singleHomo_CPU}).

The efficiency difference between C++ program with the CUDA toolkit and Python program with the PyTorch library both on GPU can be seen from Table~\ref{tab:runtime_compare_Tensor} and Table~\ref{tab:runtime_GPU}. Take the runtime of ACA as an instance. For deep homography inference on GPU, computing a $4$-point homography with Python statements costs $697.2$$\upmu$s, as shown in the penultimate row of Table~\ref{tab:runtime_compare_Tensor}. However, computing a $4$-point homography using CUDA only costs $3.11$$\upmu$s, as shown in the last column of Table~\ref{tab:runtime_GPU}. 
This $224$\textit{x} speedup demonstrates the speed advantage of directly implementing deep homography networks with the C++ CUDA toolkit.

\vspace{-2mm}
\section{Conclusion and Future Work}
\label{sec:conclusion}
In this paper, two novel homography decomposition methods with high computational efficiency and clear geometrical interpretation are presented. In particular, the kernel transformation in the SKS decomposition with only four parameters indicates the minimal expression of projective distortion between two normalized images.
The ACA decomposition further significantly improves the computational efficiency, requiring only $97$ FLOPs ($85$ FLOPs and no division operations for homographies up to a scale).
Experiments conducted on a consumer desktop computer show that our ACA algorithm can be run about \textbf{$70$M} times on CPU and \textbf{$4$G} times on GPU.
This huge computational advantage enables the proposed algorithm to be integrated into both the traditional feature-based RANSAC pipeline and the deep homography pipeline to replace their default $4$-point homography solvers. In addition, the proposed anchor points based method is a unified solution for planar primitives, which has been illustrated in the extension of SAP decomposition and the computation of 2D affine transformation.

Practical applications of other advantages, such as the direct expression of each element in homography via a polynomial of input coordinates and geometric parameterization, are worth pursuing in future work. We are also willing to explore the significance of SKS and ACA in other vision problems, such as pose estimation and image stitching.

\appendices
\renewcommand{\theequation}{\thesection.\arabic{equation}}
\section*{Appendix}

\section{Proof of Expression of Kernel Transformation}
\label{appendix:A}
\setcounter{equation}{0} 
Assume that the kernel transformation is expressed by
\begin{equation}
\mathbf{H}_{K} = 
\begin{bmatrix}
k_{11} & k_{12} & k_{13} \\
   k_{21}  & k_{22} & k_{23}  \\
 k_{31} & k_{32} & k_{33} \end{bmatrix},
\end{equation}
where $\mathbf{H}_{K}$ should satisfy
\begin{equation}
\begin{bmatrix}
k_{11} & k_{12} & k_{13} \\
   k_{21}  & k_{22} & k_{23}  \\
 k_{31} & k_{32} & k_{33} \end{bmatrix}
 \begin{bmatrix}
\mp1 \\
    0 \\
 1 \end{bmatrix} \equiv
 \begin{bmatrix}
\mp1 \\
    0 \\
 1 \end{bmatrix}.
\end{equation}

Thus we have
\begin{equation}
\begin{split}
-k_{21}+k_{23} = 0,& \quad k_{21}+k_{23} = 0, \\
k_{11}+k_{13} = k_{31}+k_{33},& \quad 
-k_{11}+k_{13} = k_{31}-k_{33}. \\
\end{split}
\end{equation}

Solving the above equations, we have
\begin{equation}
k_{21}=0, \quad k_{23} = 0, \quad 
k_{11}=k_{33}, \quad k_{13} = k_{31}.
\end{equation}

Removing the homogeneous equality (let $k_{22}=1$), the kernel transformation can be expressed by
\begin{equation}
\mathbf{H}_{K} \equiv \begin{bmatrix}
a_K & u_K & b_K \\
     & 1 &   \\
 b_K & v_K & a_K \end{bmatrix}.
 \label{equ:A_5}
 \end{equation}

\section{Failure Cases of RHO-GE}
\label{appendix:C}
In RHO-GE~\cite{Bazargani2015Fast}, since no pivoting is adopted, division by zero will occur for some specific configurations of four points on source plane. 
Still denoting the four points on source plane and their correspondences on target plane by $\{\!M_1,\!N_1\!,\!P_1\!,\!Q_1\!\}$ and   $\{\!M_2,\!N_2\!,\!P_2\!,\!Q_2\!\}$ respectively. In RHO-GE, the order of four points is fixed and the third point $P_1$ is used to be subtracted for the first time of row operation. After some row interchange operations, the first six columns of 
$\mathbf{A}_{8*9}$ in Eq.~\ref{equ:intro_3} is expressed by
\vspace{-2mm}

\begin{scriptsize}
\begin{equation}
\begin{bmatrix}
M_1.x\!-\!P_1.x & M_1.y\!-\!P_1.y & 1 & 0 & 0 & 0 \\
N_1.x\!-\!P_1.x & N_1.y\!-\!P_1.y & 1 & 0 & 0 & 0 \\
P_1.x & P_1.y & 1 & 0 & 0 & 0 \\ 
Q_1.x\!-\!P_1.x & Q_1.y\!-\!P_1.y & 1 & 0 & 0 & 0 \\
0 & 0 & 0 & M_1.x\!-\!P_1.x & M_1.y\!-\!P_1.y & 1 \\
0 & 0 & 0 & N_1.x\!-\!P_1.x & N_1.y\!-\!P_1.y & 1 \\
0 & 0 & 0 & P_1.x & P_1.y & 1  \\
0 & 0 & 0 & Q_1.x\!-\!P_1.x & Q_1.y\!-\!P_1.y & 1 \\
\end{bmatrix}.
\label{equ:D_1}
\end{equation}
\end{scriptsize}

\vspace{3mm}
In the subsequent row reduction operations, two fixed items will be selected as the divisors, which are expressed by
\vspace{-2mm}

\begin{scriptsize}
\begin{equation}
\begin{split}
d_1 &= M_1.x\!-\!P_1.x, \\
d_2 &= (M_1.x\!-\!P_1.x)*(N_1.y\!-\!P_1.y) - (M_1.y\!-\!P_1.y)*(N_1.x\!-\!P_1.x).
\end{split}
\label{equ:D_2}
\end{equation}
\end{scriptsize}

\vspace{-3mm}
Obviously, the first point $M_1$ and the third point $P_1$ on the source plane cannot be on a vertical line, otherwise the divisor $d_1$ will be zero. 
However, the implementation of RHO-GE in OpenCV (the fastest competitor among the four previous methods in our C++ experiments), does not contain program statements (which would degrade its speed performance) to exclude such failure cases.
For a rectangle on source plane, among $24$ orders of $4$ vertices, $8$ orders will divide by $0$. Therefore, a manual ordering of four vertices is done by us to avoid this failure case in our deep learning experiments. For the other divisor $d_2$, although it cannot be zero unless the first three points are collinear or two of them are same (both of which are degenerate configurations), $d_2$ may cause a decrease in numerical precision when it is close to zero.

{
\small
\bibliographystyle{ieee}

}

\end{document}